%% file: iclr2027_conference.tex
\documentclass{article} \usepackage{iclr2027_conference,times}

\input{math_commands.tex}

\usepackage{multirow}
\usepackage{hyperref}
\usepackage{url}
\usepackage{booktabs}
\usepackage{graphicx}
\usepackage{array}
\usepackage{capt-of}
\usepackage{flafter}
\usepackage{wrapfig}
\usepackage{microtype}
\usepackage[section]{placeins}
\usepackage{hyperref}
\hypersetup{
	colorlinks=false,
	citebordercolor={0 1 0},
	pdfborder={0 0 1}
}
\title{From Complex Histories to Simple Predictors: Predictive Dimension in Dynamic Graph Learning}

	\author{
			Hantong Feng$^{1,4}$\\
			\And
			Duxin Chen$^{2,4}$\\
			\And
			Wenwu Yu$^{2,3,4*}$\\
			\AND
			$^1$School of Cyber Science and Engineering, Southeast University.\\
			$^2$School of Mathematics, Southeast University.\\
			$^3$Frontiers Science Center for Mobile Information Communication and Security.\\
		}

\iclrfinalcopy

\begin{document}

\maketitle
\lhead{}

\begin{abstract}
Existing continuous-time graph learning methods rely heavily on complex neural architectures to encode ever-expanding interaction histories. However, the future conditional distribution induced by complex histories can often be characterized by a low-dimensional predictive state manifold, implying that only a few predictive degrees of freedom are essential for future interactions. We therefore argue that continuous-time graph learning should identify low-dimensional states sufficient for future prediction rather than reconstruct the full history. To make this perspective measurable, we define and independently estimate the intrinsic predictive dimension, which quantifies the effective predictive degrees of freedom required by a given task. Specifically, we remove the history-independent marginal from the history-conditioned future distribution, decompose the resulting predictive residual operator, and estimate the minimum effective rank needed to preserve the predictive power of the full history. Experiments across datasets show that model performance saturation is largely determined by the intrinsic predictive dimension. Based on this finding, we propose PreDyG, a lightweight model whose capacity is matched to the intrinsic predictive dimension. PreDyG achieves superior predictive performance with only 0.07\% of the parameters of a representative complex baseline.
\end{abstract}

\section{Introduction}

Continuous-time graph learning aims to predict future interactions from expanding timestamped event streams \cite{dynsurvey2020}. Existing methods mostly treat interaction history as the primary modeling object, representing it through node memory, temporal message passing, recurrent updates, or Transformer encoders \cite{ctdg_memory,ctdg_message_passing,ctdg_transformer}. Recent lightweight architectures suggest that such complex modeling mechanisms are not always necessary \cite{graphmixer2023,freedyg2024}. However, simplifying the architecture alone does not answer a more fundamental question: for a given prediction task, how many predictive degrees of freedom must a model actually retain?

The key observation is that observational complexity is not the same as predictive complexity. Different histories may induce similar conditional distributions over future interactions and are therefore equivalent in the predictive sense \cite{predictive_state,predictive_dimension}, as illustrated in Figure~1a. We find that these history-conditioned future distributions concentrate on a low-dimensional predictive-state manifold, meaning that only a small number of predictive degrees of freedom truly determine the future. This finding reframes continuous-time graph learning from reconstructing the full history to identifying a low-dimensional state that is sufficient for future prediction.

We formalize this structure through the intrinsic predictive dimension. Given a prediction target and a tolerance level, the intrinsic predictive dimension is defined as the minimum predictive-state dimension required to retain a prescribed fraction of the achievable future-predictive gain. To estimate it independently of the final neural model, we first subtract the no-history marginal distribution from the history-conditioned future distribution, and then perform a low-rank decomposition of the resulting predictive residual operator. We then select, on held-out data, the minimum effective rank that approximately preserves the predictive ability of the full history. This separation makes predictive dimension a testable data-side quantity rather than an architecture-dependent hyperparameter. It also allows the predictive degrees of freedom required by a task to be measured before architecture design and independently of the final neural model.

Across diverse real continuous-time networks from social, communication, and mobility domains, we observe consistent low-dimensional predictive structure. The estimated predictive dimension is positively aligned with the saturation point of predictive gain and parameter budget, provides a more informative capacity ordering than node or event counts. After normalizing model dimension by the estimated predictive dimension, capacity curves become more consistent across datasets: model performance typically enters a plateau when the state dimension approaches the predictive dimension, whereas under-dimensioned models recover substantially less predictive gain. Figure~1b shows that, despite large differences in network scale, event count, and raw performance, predictive gains exhibit similar rising and saturation trends when normalized by intrinsic predictive dimension.

We further examine whether this relationship arises from genuine history--future dependencies or merely reflects statistical differences associated with dataset identity. To this end, we construct intervention experiments that preserve selected marginal statistics while separately disrupting temporal order, source-node identity, destination-node identity, or event pairing. These interventions induce coherent changes in the estimated predictive dimension and the independently measured saturation dimension, while also revealing dataset-specific responses to temporal and structural dependencies. This evidence argues against interpreting predictive dimension as a proxy for graph size, event count, dataset identity, or generic compressibility. Rather, it captures the effective predictive directions through which history constrains future interactions.
Based on this empirical regularity, we propose the Predictive Dimension Principle: within a fixed architecture family, model-state capacity should match the intrinsic predictive dimension of the interaction process, rather than the scale of the observed history. Guided by this principle, we design \textbf{Pre}dictive dimension-Guided \textbf{Dy}namic \textbf{G}raph Learning (\textbf{PreDyG}), an extremely compact architecture that uses low-dimensional states to model effective predictive degrees of freedom. Its compactness does not come from post-hoc pruning or model compression, but from allocating model capacity according to a data-side estimate of predictive complexity. Experiments compare the dimension-guided model with under-capacity models, over-capacity models, and conventional temporal graph architectures in terms of predictive accuracy, parameter count, training cost, and inference time. Our contributions are summarized as follows:
\begin{itemize}
	\item We formalize intrinsic predictive dimension for continuous-time interaction streams and develop an estimator independent of the final neural model. This makes the predictive degrees of freedom required by a given task measurable before architecture design.
		\item We establish an empirical relationship between task-aligned predictive dimension and capacity saturation across diverse temporal networks. Predictive dimension indicates where additional capacity yields diminishing predictive gains, and experiments further connect this relationship to history–future dependencies.
		\item Guided by the Predictive Dimension Principle, we develop PreDyG, a compact temporal graph model. It achieves the best performance among the compared methods on most evaluated datasets across diverse domains, using only a significantly smaller number of trainable parameters compared to the baseline.
	\end{itemize}
\section{Related Work}

\paragraph{Continuous-time graph learning.}
Continuous-time graph models encode timestamped histories through recurrent memory and message passing \cite{jodie2019,ctdg_message_passing,ctdg_memory}, temporal neighborhoods and walks \cite{tgat2020,cawn2021,neurtw2022}, or mixing, Transformer, frequency-domain, and state-space operators \cite{tcl2021,graphmixer2023,ctdg_transformer,freedyg2024,ctan2024,dygmamba2025}. Complementary work studies model expressivity and execution efficiency \cite{pint2022,nat2022,tgl2022,tpnet2024}. These studies mainly ask how history should be represented or processed. GraphMixer and CRAFT \cite{graphmixer2023,yi2026future} show that simplified temporal-graph architectures can remain competitive, but neither provides a data-side criterion for determining when a compact predictive state is sufficient.

\paragraph{Temporal graph complexity.}
Predictive-state representations define state by the future distributions it preserves \cite{predictive_state}, while predictive dimension measures the past features needed for informative prediction \cite{predictive_dimension}. Low-dimensional network geometry and shortest compression length provide complementary notions of structural complexity and predictability \cite{ultralow2022,shortestcompression2020}; temporal null models isolate genuine temporal dependence \cite{temporalnull2022}. However, these lines neither estimate prediction-sufficient dimension for continuous-time graph events nor connect it to capacity saturation. We address both gaps by independently estimating predictive dimension and using it to guide model capacity.

\section{Problem Formulation}
\label{sec:problem}

\paragraph{Continuous-time interaction.}
Let $\mathcal{V}$ be the node set and $\mathcal{E}=\{e_k\}_{k=1}^{M}$ a chronologically ordered event stream, where $e_k=(u_k,v_k,t_k,\vx_k)$ denotes an interaction from source $u_k\in\mathcal{V}$ to destination $v_k\in\mathcal{V}$ at time $t_k$, optionally with event features $\vx_k$. The history immediately before $t_k$ is $\mathcal{H}_k=\{e_j:t_j<t_k\}$. Given a query $q_k$ (e.g., a source and timestamp), temporal link prediction estimates a future variable $Y_{k,h}$, such as the destination of the next interaction or the existence of an edge within horizon $h$.

\paragraph{Intrinsic predictive dimension.}
A state map $g_d$ compresses the history into $Z_k=g_d(\mathcal{H}_k,q_k)\in\mathbb{R}^d$. Histories $\mathcal{H}$ and $\mathcal{H}'$ are predictively equivalent for target $Y_{k,h}$ when
\begin{equation}
 p(Y_{k,h}\mid \mathcal{H},q_k)=p(Y_{k,h}\mid \mathcal{H}',q_k).
 \label{eq:predictive-equivalence}
\end{equation}
where $p(\cdot\mid\cdot)$ denotes a conditional distribution, $Y_{k,h}$ is the prediction target at horizon $h$ for query $k$, $q_k$ is the query information, and $\mathcal{H}$ and $\mathcal{H}'$ are two candidate histories.
The equivalence classes can be far fewer than the observed histories. Accordingly, graph size and history length are not, by themselves, measures of predictive structural complexity.
For a fixed prediction target, horizon, representation family, loss $\ell$, and tolerance $\epsilon$, define the best attainable risk at dimension $d$ as
\begin{equation}
 \mathcal{R}_h(d)=\inf_{g_d,a}\;\mathbb{E}\!\left[\ell\!\left(a(g_d(\mathcal{H}_k,q_k)),Y_{k,h}\right)\right]
 \label{eq:risk-d}
\end{equation}
where $g_d$ maps the history and query to a state in $\mathbb{R}^d$, $a$ maps that state to a prediction, $\ell$ is the task loss, and the expectation is over query--target pairs from the interaction process. Let $\mathcal{R}_0$ be the no-history risk and $\mathcal{R}_h^\star=\inf_d\mathcal{R}_h(d)$ the unrestricted reference risk. We define
\begin{equation}
 d_{\mathrm{pred}}(\epsilon,h)=
 \min\!\left\{d:\mathcal{R}_h(d)\leq
 \mathcal{R}_h^\star+\epsilon\big(\mathcal{R}_0-\mathcal{R}_h^\star\big)\right\}.
 \label{eq:dpred}
\end{equation}
where $\epsilon\in[0,1]$ is the tolerated fraction of the predictive gain not recovered by the state, and $h$ specifies the future horizon.
Thus $d_{\mathrm{pred}}$ measures the state dimension needed to recover a $1-\epsilon$ fraction of the predictive gain supplied by history. It measures predictive structure, not irreducible uncertainty and not a globally minimum neural-network parameter count. Its relationship to neural parameter count is therefore empirical and is evaluated within a fixed model family and training protocol.

\section{Methodology}
\label{sec:method}

Our methodology separates measurement from model construction. We first define a model-independent estimator of predictive dimension, then state the empirical capacity law it is used to test. Only after establishing this data-side quantity do we introduce the Predictive Dimension Principle and the PreDyG architecture.
\subsection{Independent Predictive Dimension Estimator}
\label{sec:estimator}
\begin{figure*}[t]
	\centering
	\includegraphics[scale=0.47]{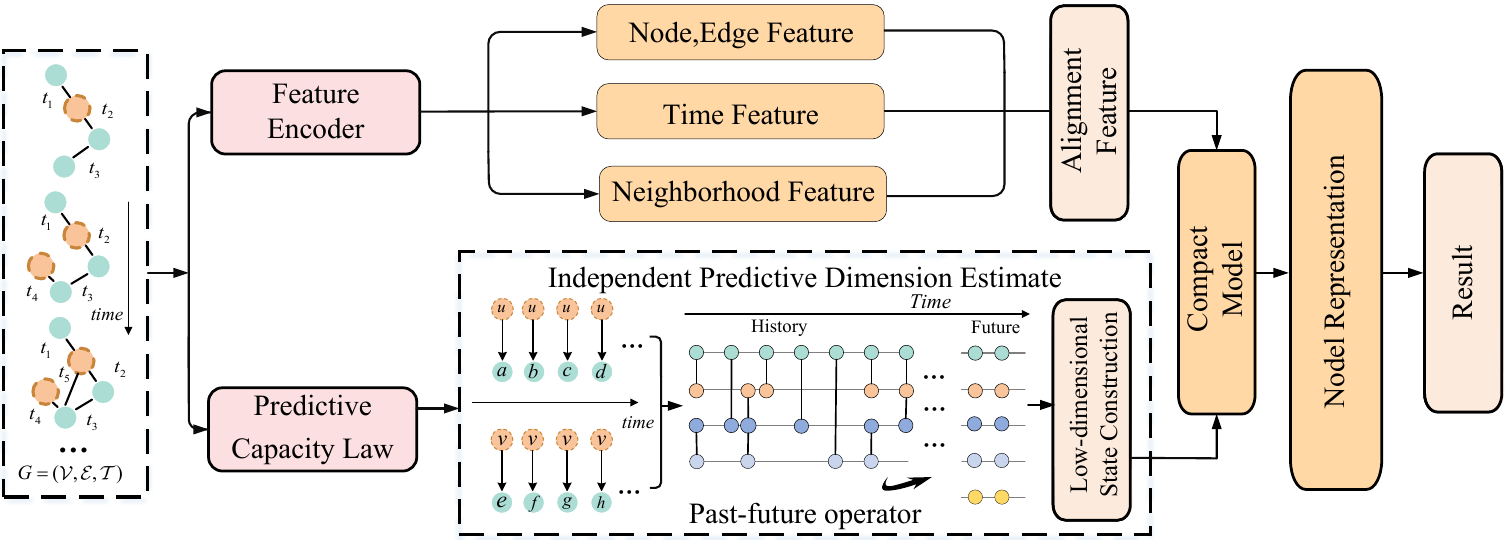}
	\caption{Overview of PreDyG, the encoder guided by intrinsic predictive dimension.}
	\label{4}
\end{figure*}
For finite event streams, estimating Equation~\ref{eq:dpred} over unrestricted histories is ill posed. We construct a task-aligned future-distribution representation with ambient dimension \(D_{\mathrm{amb}}\). Source and destination identities are grouped using a fixed frequency-based rule, and the resulting representation is held identical across datasets. For event $k$, the context is
\begin{equation}
 H_k=\big(b_s(u_k), b_d(v_{\operatorname{prev}(u_k)})\big),
 \qquad F_k=b_d(v_k),
 \label{eq:probe}
\end{equation}
where $b_s$ and $b_d$ are the fixed source- and destination-frequency binning maps, $v_{\operatorname{prev}(u_k)}$ is the most recent destination previously contacted by $u_k$, $H_k$ is the discrete past context, and $F_k$ is the one-step future label. This estimator is held identical across domains; it is not tuned to PreDyG.

\paragraph{Past--future operator.}
We estimate $Q$ from event counts with additive $0.5$ smoothing and center its history-dependent component:
\begin{equation}
 C_{hf}=\sqrt{p_h}\big(Q_{hf}-\pi_f\big).
 \label{eq:centered-operator}
\end{equation}
where $h$ and $f$ index values of $H$ and $F$, $Q_{hf}=p(F=f\mid H=h)$ is a conditional probability, $p_h=p(H=h)$ is the context marginal, $\pi_f=\sum_h p_hQ_{hf}$ is the no-history future marginal, and $C$ is the centered, marginal-weighted past--future operator. The marginal $\pi$ is the one-dimensional no-history component. For a candidate dimension $d$, we retain $d-1$ orthogonal predictive coordinates of $C$, computed by truncated singular-value decomposition, reconstruct $Q^{(d)}$, clip negative entries, and normalize rows to the probability simplex. The singular values are not used directly as a complexity statistic; the predictive dimension is selected by held-out Brier risk.

\paragraph{Predictive dimension selection.}
We evaluate a logarithmically spaced rank grid up to $D_{\mathrm{amb}}$. The first $40\%$ of events fit the operator and the next $20\%$ form the selection split. With $R^*=\min_dR(d)$ and $\epsilon=0.05$, the predictive-dimension estimate is
\begin{equation}
 \widehat d_{\mathrm{pred}}=\min\{d:R(d)\leq R^*+0.05[R(1)-R^*]\}.
 \label{eq:prd}
\end{equation}
where $R(d)$ is selection-split Brier risk at dimension $d$, $R^*=\min_dR(d)$ is the best risk among candidate dimensions, and $R(1)$ is the no-history reference risk; $0.05$ fixes the tolerated loss of historical predictive gain.
The first $70\%$ are then refit and the last $30\%$ independently determine the held-out predictive-rank estimate $d_{\mathrm{hold}}$ using the same $5\%$ rule. This chronology lets $\widehat d_{\mathrm{pred}}$ predict, rather than define, $d_{\mathrm{hold}}$.

\subsection{Empirical Capacity Law}
\label{sec:law}
To compare predictive saturation across datasets, we normalize the history-dependent risk reduction as
\begin{equation}
	G_D(d)=
	\frac{R_D(1)-R_D(d)}
	{R_D(1)-\min_j R_D(j)}.
	\label{eq:normalized-gain}
\end{equation}
where $R_D(d)$ is the held-out risk at predictive-state dimension $d$, $R_D(1)$ is the no-history risk, and $j$ ranges over the candidate dimensions. When historical gain is positive, $G_D(d)=1$ denotes the best gain observed in the dimension scan. A valid predictive-dimension estimate should organize these curves, with predictive gain approaching its plateau as $d/\widehat d_{\mathrm{pred}}$ approaches one.

\subsection{Predictive Dimension Principle}
\label{sec:principle}

The empirical capacity law motivates the following design rule:
\begin{quote}
	\emph{For a fixed prediction target, horizon, and tolerance, allocate model-state capacity according to the independently estimated intrinsic predictive dimension of the interaction process, rather than according to its number of nodes, events, or historical observations.}
\end{quote}

This principle treats predictive dimension as a data-side scale for selecting a sufficient trainable-parameter budget. Within a fixed PreDyG architecture family, trainable capacity is jointly distributed across the predictive state, modality projections, fusion module, and task-specific readout. We therefore use $\widehat d_{\mathrm{pred}}$ to identify the parameter regime in which predictive performance saturates.
\subsection{PreDyG: A Predictive-Dimension Guided Encoder}
\label{sec:PreDyG}
we proposed PreDyG which implements the principle by parameterizing a low-dimensional predictive state, using explicit interaction statistics to expose recurring structure, and restricting learnable capacity to the components needed for future prediction.
\paragraph{Interaction-history features.}
For a query edge $(u,v,t)$, retrieves the $L$ most recent interactions of each endpoint. For neighbor $w_j$ at time $t_j$, four inputs are constructed: its node feature, the edge feature, a fixed sinusoidal encoding of $t-t_j$, and a neighborhood-interaction-frequency (NIF) statistic. For endpoint pair $(u,v)$, the latter records how often $w_j$ appears in the local histories of $u$ and $v$,
\begin{equation}
 \vc_{u,v}(w_j)=\left[\sum_{a\in\mathcal{N}_L(u,t)}\mathbf{1}[a=w_j],
 \sum_{a\in\mathcal{N}_L(v,t)}\mathbf{1}[a=w_j]\right],
 \label{eq:nif}
\end{equation}
where $\mathcal{N}_L(x,t)$ is the multiset of neighbors in endpoint $x$'s $L$ most recent pre-$t$ interactions, $a$ indexes a neighbor occurrence, $\mathbf{1}[\cdot]$ is the indicator function, and $\vc_{u,v}(w_j)\in\mathbb{R}^2$ is the NIF count vector for neighbor $w_j$.
This statistic exposes repeated and cross-neighborhood interaction structure without a learned node-sized memory table.

\paragraph{Low-dimensional state construction.}
Each modality is projected to a width-$m$ channel and fused:
\begin{equation}
 \vz_j=W_r\big[W_x\vx_{w_j}\,\Vert\,W_e\ve_j\,\Vert\,
 W_t\boldsymbol{\tau}(t-t_j)\,\Vert\,W_n\phi(\vc_{u,v}(w_j))\big],
 \label{eq:token}
\end{equation}
where $\vx_{w_j}$ and $\ve_j$ are the neighbor-node and event features, $\boldsymbol{\tau}$ is the fixed sinusoidal elapsed-time encoding, $\phi$ is the NIF encoder, $W_x,W_e,W_t,W_n$ are modality projections, $W_r$ is the fusion projection, $\Vert$ denotes concatenation, and $\vz_j$ is the fused historical interaction.
The compact instantiation combines three summaries of the historical interaction representations, namely their mean, coordinatewise maximum, and most recent representation, through a single projection:
\begin{equation}
 \vh_u(t)=W_p\big[\operatorname{mean}_j\vz_j\,\Vert\,
 \operatorname{max}_j\vz_j\,\Vert\,\vz_{L}\big],
 \label{eq:pool}
\end{equation}
where $\vh_u(t)$ is the predictive state of $u$, $W_p$ is the pooling projection, the mean and maximum are computed coordinatewise over the $L$ historical interaction representations, and $\vz_L$ denotes the most recent representation.
The same encoder produces $\vh_v(t)$, and a linear pair readout gives
\begin{equation}
\hat p(u,v,t)=\sigma(\vw_s^\top[\vh_u(t)\Vert\vh_v(t)]),
\label{eq:score}
\end{equation}
where $\hat p(u,v,t)$ is the predicted interaction probability, $\sigma$ is the logistic sigmoid, and $\vw_s$ is the learned pair-readout vector.
Training uses binary cross-entropy with sampled negative destinations. The model has no recurrent node memory, temporal message-passing stack, or Transformer stack.

\paragraph{Dimension-guided capacity.}
For a fixed PreDyG architecture family, we use the task-aligned predictive dimension to identify a sufficient trainable-parameter budget. Specifically, we select the smallest configuration whose validation performance reaches at least $95\%$ of the best result in the capacity scan. Predictive dimension therefore guides the total parameter budget, it is distributed across the predictive state, modality projections, fusion module, and readout. In general, the trainable parameter count scales as $O(m(p_x+p_e+p_t+p_n)+m^2+p_n^2)$, where $m$ is the predictive-state width and $p_x,p_e,p_t,p_n$ are the input widths of the node, edge, time, and NIF channels.

\section{Experiments}
\label{sec:experiments}

\input{iclr2027_experiments_revised.tex}

\section{Conclusion}
This work estimates the prediction degrees independently of the final neural network model based on the concept of intrinsic prediction dimension. In various continuous-time networks, the task-aligned predictive dimension provides a more informative ordering of sufficient parameter budgets than graph scale. These findings prompt the proposal of the "prediction dimension principle": within a fixed architecture family, the capacity of the model state should match the degrees of freedom of prediction relevance, rather than the observed historical quantities. PreDyG implements this principle through a lightweight prediction state encoder, achieving performance comparable to or exceeding that of larger-scale models in different evaluation domains. Its compactness stems from the measured prediction structure, whose parameter regime is selected using the predictive-dimension analysis.

\subsection*{AI use statement}
In this work, we used generative AI tools for language polishing, grammar correction, and wording refinement of the manuscript. We have not used generative AI tools for generating research ideas, formulating hypotheses, conducting experiments, and other required disclosure tasks are not applicable to this work. Additionally, we used generative AI tools for improving readability and clarity of presentation. We have reviewed all AI-assisted work. All AI-assisted edits were checked and revised by the authors to ensure technical correctness, consistency with the experimental results, and absence of unsupported claims. We take responsibility for the final content of this work, including text, claims or artifacts produced with the aid of generative AI.

\bibliography{phydyg_refs}
\bibliographystyle{iclr2027_conference}

\input{iclr2027_supplementary.tex}

\end{document}

%% file: math_commands.tex
\usepackage{amsmath,amsfonts,bm}

\def\eqref#1{equation~\ref{#1}}

\def\1{\bm{1}}

\def\vc{{\bm{c}}}

\def\ve{{\bm{e}}}

\def\vh{{\bm{h}}}

\def\vw{{\bm{w}}}
\def\vx{{\bm{x}}}

\def\vz{{\bm{z}}}

\DeclareMathAlphabet{\mathsfit}{\encodingdefault}{\sfdefault}{m}{sl}
\SetMathAlphabet{\mathsfit}{bold}{\encodingdefault}{\sfdefault}{bx}{n}

%% file: iclr2027_experiments_revised.tex
We organize the experiments in two stages. Stage~I estimates task-aligned predictive dimension and tests whether it organizes predictive-gain saturation across datasets. Stage~II examines whether this data-side scale can guide the sufficient parameter budget of PreDyG, and then evaluates its predictive performance and efficiency.
\textbf{Data-side measurement.}
We study 12 continuous-time interaction streams spanning communication, proximity, politics, social behavior, recommender systems, and mobility,
using at most the first 200,000 events of each stream. All datasets share the same prediction target and history representation, with Brier risk and a $5\%$ tolerance. Predictive dimension is estimated on chronological data splits before neural-model training. Estimation details are provided in Appendix~\ref{app:predictive-dimension-settings}; comparisons with scale and compression statistics are reported in Appendix~\ref{app:statistic-comparison}.
\textbf{Model-side validation.}
Dynamic link prediction uses chronological $70\%/15\%/15\%$ train/validation/test splits, with evaluation in transductive and inductive settings. We report average precision (AP) as mean $\pm$ standard deviation across repeated runs. We conducted experiments on 8 widely diverse datasets, and also set up 4 large TGB datasets following their official data splitting method and MRR evaluation protocol~\cite{tgb2023}. For the capacity analysis, we define the sufficient parameter budget $P_{0.05}(D)$ as the smallest trainable parameter count whose validation AP reaches at least $95\%$ of the best value observed in the capacity scan. Datasets and baseline methods are described in Appendices~\ref{app:ComprehensiveDatasetCollection} and~\ref{app:BaselineAlgorithmLandscape}, respectively.

\subsection{Predictive Dimension Analysis}
\label{sec:stage_1}
\begin{wrapfigure}[25]{r}{0.41\textwidth}
		\centering
	\includegraphics[width=\linewidth]{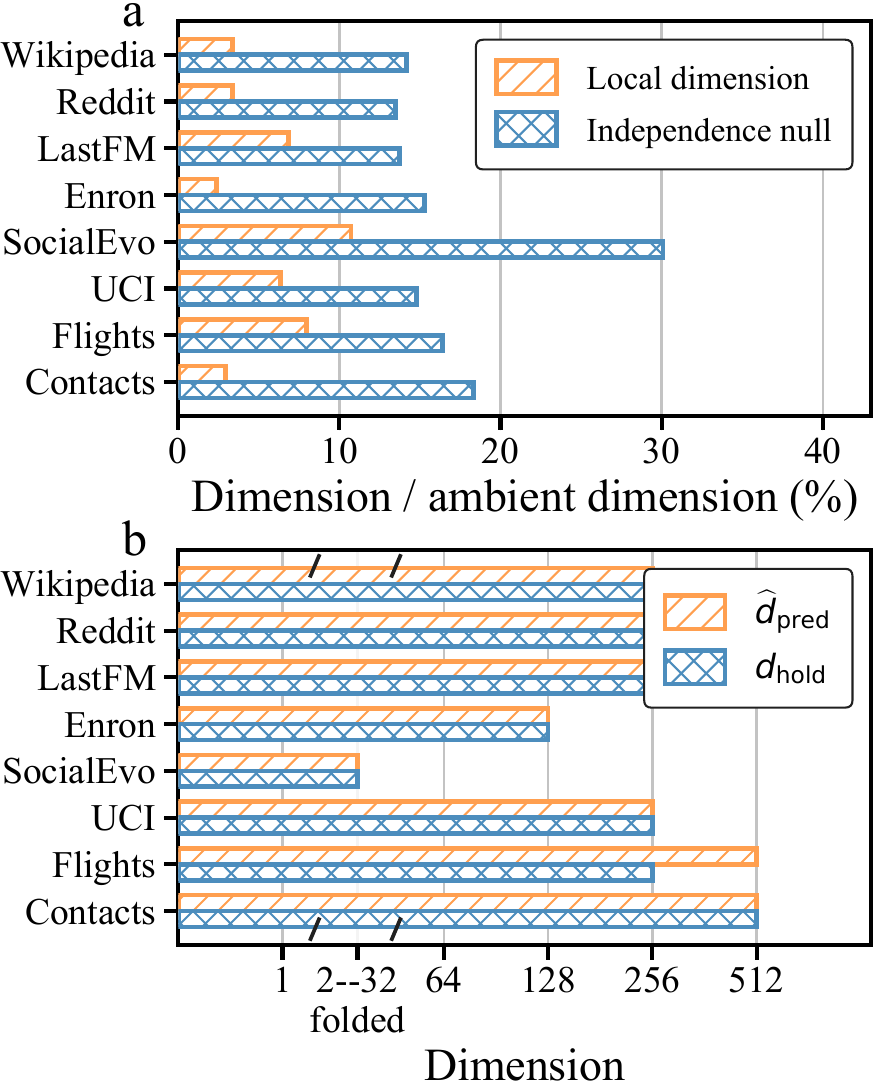}
	\caption{(a) History-conditioned future distributions are lower-dimensional than the independence null. (b) The estimated predictive dimension $\widehat d_{\mathrm{pred}}$ aligns with the held-out saturation dimension $d_{\mathrm{hold}}$, marking the onset of capacity saturation.}
	\label{fig:local-geometry}
	\end{wrapfigure}
\subsubsection{Predictive Dimension as a Capacity Scale}
We first test whether the future distributions induced by interaction histories occupy a compact predictive geometry. For each history context, we estimate its conditional future distribution and measure the local intrinsic dimension of these predictive fingerprints. As a reference, we use an independence null that preserves observation counts and future-event marginals while removing history--future dependence. This comparison asks whether the observed low-dimensionality reflects genuine predictive structure rather than marginal event statistics. Figure~\ref{fig:local-geometry}(a) shows that the estimated local dimension is only a small fraction of the ambient future-distribution dimension and is consistently below the independence-null reference. This indicates that histories do not scatter future distributions arbitrarily in the ambient space; instead, they induce a compact set of reusable predictive directions. 
We next ask whether this low-dimensional structure is relevant to model capacity. Figure~\ref{fig:local-geometry}(b) compares the independently estimated predictive dimension \(\widehat d_{\mathrm{pred}}\) with the held-out dimension \(d_{\mathrm{hold}}\). The two estimates coincide on most streams and remain close on the discrete dimension grid, suggesting that \(\widehat d_{\mathrm{pred}}\) identifies the capacity scale at which additional predictive-state dimensions provide diminishing returns.

\subsubsection{Predictive Dimension as a Dependency Measure}
To test whether predictive dimension reflects genuine history--future dependencies rather than dataset identity or marginal event statistics, we construct dependency interventions that preserve event count while perturbing temporal order, destination identity, source identity, or source--destination pairs. We apply these interventions to first 80,000-event of Reddit, SocialEvo, and Contacts, and to the complete UCI stream. These streams span different predictive-dimension regimes and dependency profiles, while the first 80,000-event keeps the intervention budget comparable across larger datasets; UCI is used in full because it is shorter. For each stochastic intervention, we report the median over three independently shuffled realizations.

As Table~\ref{tab:intervention} shows, the two independently selected dimensions move concordantly across the 20 dataset--intervention pairs, but the dependency carrying predictive dimension differs by dataset. Perturbing destination or source identity reduces all four streams to, or close to, the one-dimensional floor. Temporal and pair perturbations instead produce dataset-specific responses: UCI falls from $256/256$ to $1/1$ under either perturbation, SocialEvo retains $32/16$ after temporal shuffling and falls to $16/16$ after pair shuffling, whereas Reddit and Contacts retain their original dimensions. Thus, predictive dimension responds selectively to the temporal and structural dependencies that constrain future interactions, rather than to event count, arbitrary randomization, or a fixed dataset label.

Together, these data-side analyses show that, for a fixed prediction target, tolerance, and estimator family, predictive gain generally saturates when the available state dimension reaches the independently estimated predictive dimension. The above results all connect $\widehat d_{\mathrm{pred}}$ to future-predictive degrees of freedom rather than observation scale or dataset identity.
\begin{table*}[t]
	\caption{Dependency interventions. Each cell reports
		$\widehat d_{\mathrm{pred}}/d_{\mathrm{hold}}$, followed by the percentage
		of the original predictive dimension retained.}
	\label{tab:intervention}
	\centering
	\footnotesize
	\setlength{\tabcolsep}{5.0pt}
	\renewcommand{\arraystretch}{1.06}
	\begin{tabular}{@{}lccccc@{}}
		\toprule
		Dataset & Original & Time/order & Destination & Source &
		Source--destination pair \\
		\midrule
		Reddit
		& $256/256$ (100\%) & $256/256$ (100\%) & $1/1$ ($<1\%$)
		& $1/1$ ($<1\%$) & $256/256$ (100\%) \\
		SocialEvo
		& $32/32$ (100\%) & $32/16$ (100\%) & $1/1$ (3\%)
		& $1/1$ (3\%) & $16/16$ (50\%) \\
		UCI
		& $256/256$ (100\%) & $1/1$ ($<1\%$) & $1/1$ ($<1\%$)
		& $1/1$ ($<1\%$) & $1/1$ ($<1\%$) \\
		Contacts
		& $512/512$ (100\%) & $512/512$ (100\%) & $1/1$ ($<1\%$)
		& $4/1$ (1\%) & $512/512$ (100\%) \\
		\bottomrule
	\end{tabular}
\end{table*}

\subsection{From Predictive Dimension to Model Design}
\label{sec:stage_2}

\subsubsection{Predictive Dimension Guides Capacity Selection}

\begin{wrapfigure}[25]{r}{0.41\textwidth}
\vspace{-0.7\baselineskip}
\centering
\includegraphics[width=\linewidth]{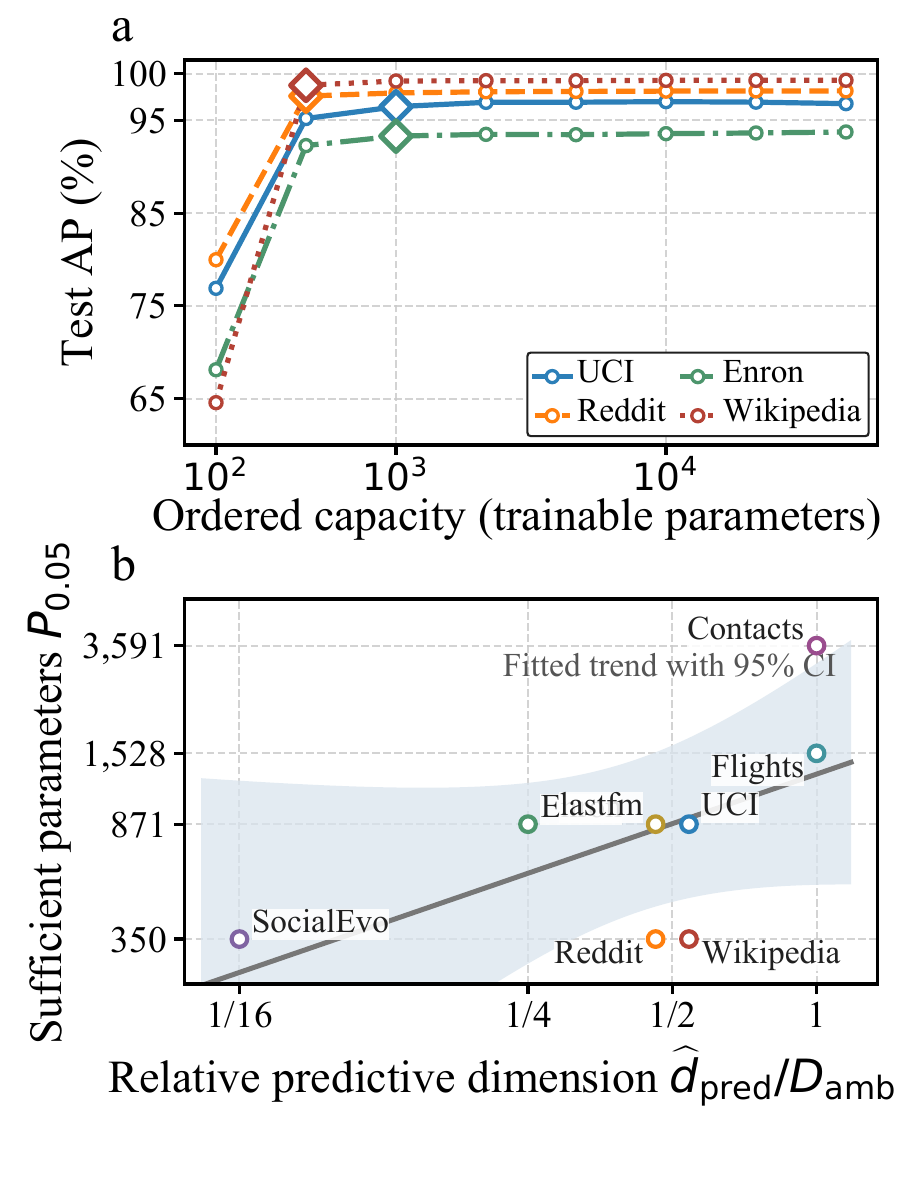}
\caption{(a) Test AP across ordered capacity configurations; diamonds mark $P_{0.05}$. (b) $P_{0.05}$ versus $\widehat d_{\mathrm{pred}}/D_{\mathrm{amb}}$; the line and shading show the fitted trend and its $95\%$ confidence interval.}
\label{fig:dimension-match}
\vspace{-0.7\baselineskip}
\end{wrapfigure}
We next test whether predictive dimension can guide a practical capacity choice. 
For each dataset, we keep the prediction task, data split, and training procedure fixed, and evaluate an ordered family of PreDyG configurations. We define the sufficient parameter budget as the smallest configuration that reaches at least $95\%$ of the best value observed in the capacity scan.
Figure~\ref{fig:dimension-match}(a) reveals a consistent transition from under-capacity to saturation across four representative datasets. Test AP improves sharply in the low-capacity regime but changes only marginally beyond the validation-selected $P_{0.05}$. The selected configuration therefore captures the transition from under-capacity to diminishing returns, rather than simply favoring the smallest model.
Figure~\ref{fig:dimension-match}(b) extends this comparison to eight datasets. The sufficient parameter budget generally increases with relative predictive dimension: datasets with limited predictive structure saturate with smaller models, while those with richer predictive structure require more parameters. The relationship is not an exact conversion from dimension to parameter count, but it provides a data-dependent capacity scale before selecting the final architecture.

This guidance also translates into end-to-end efficiency. The selected PreDyG configuration uses average about $1300\times$ fewer trainable parameters than the representative baseline while maintaining comparable transductive and new-node prediction performance. Predictive dimension therefore turns the data-side law into an actionable model-selection principle.

							\subsubsection{Efficiency Analysis}

\begin{figure*}
	\centering
	\includegraphics[width=0.27\linewidth]{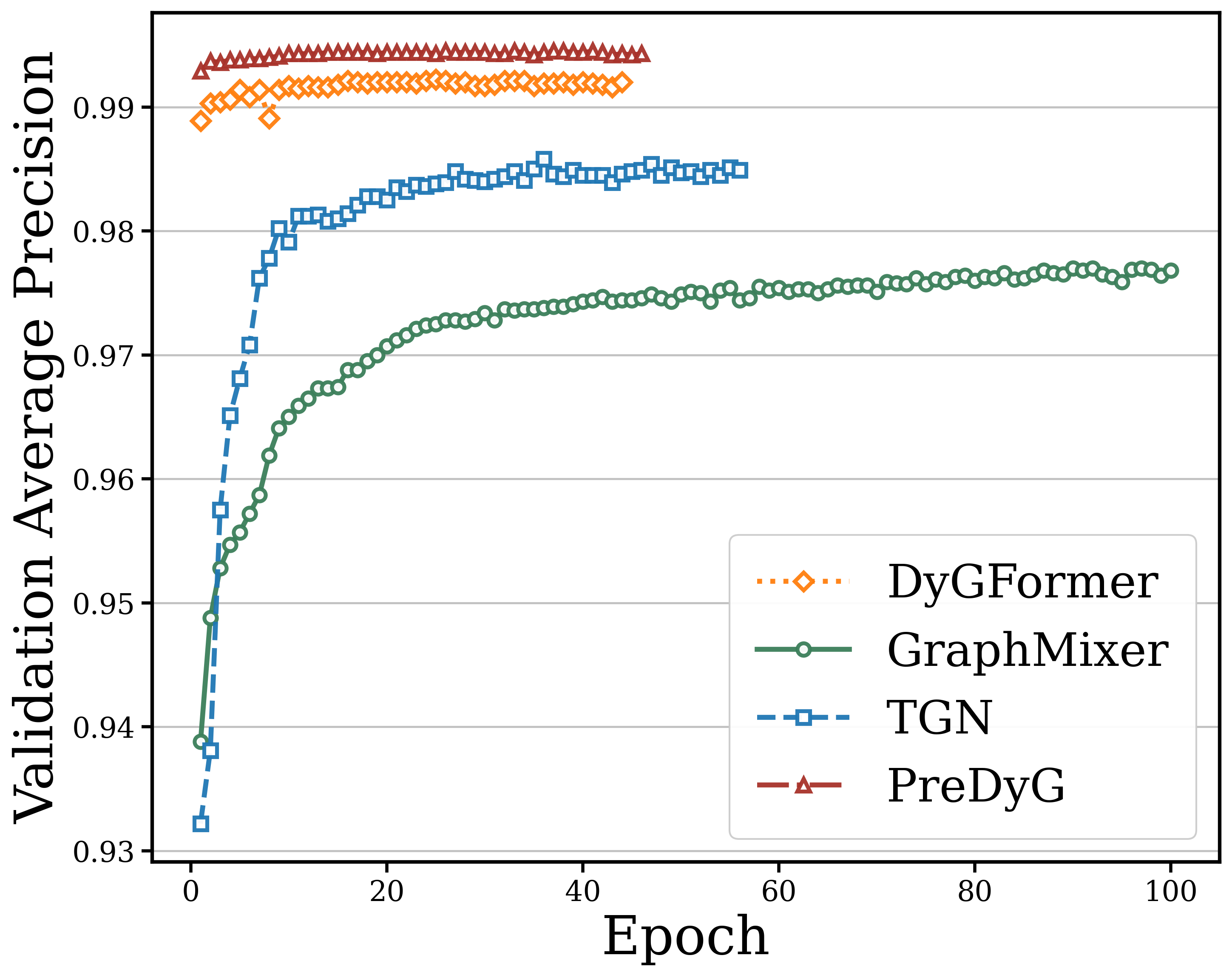}
	\includegraphics[width=0.27\linewidth]{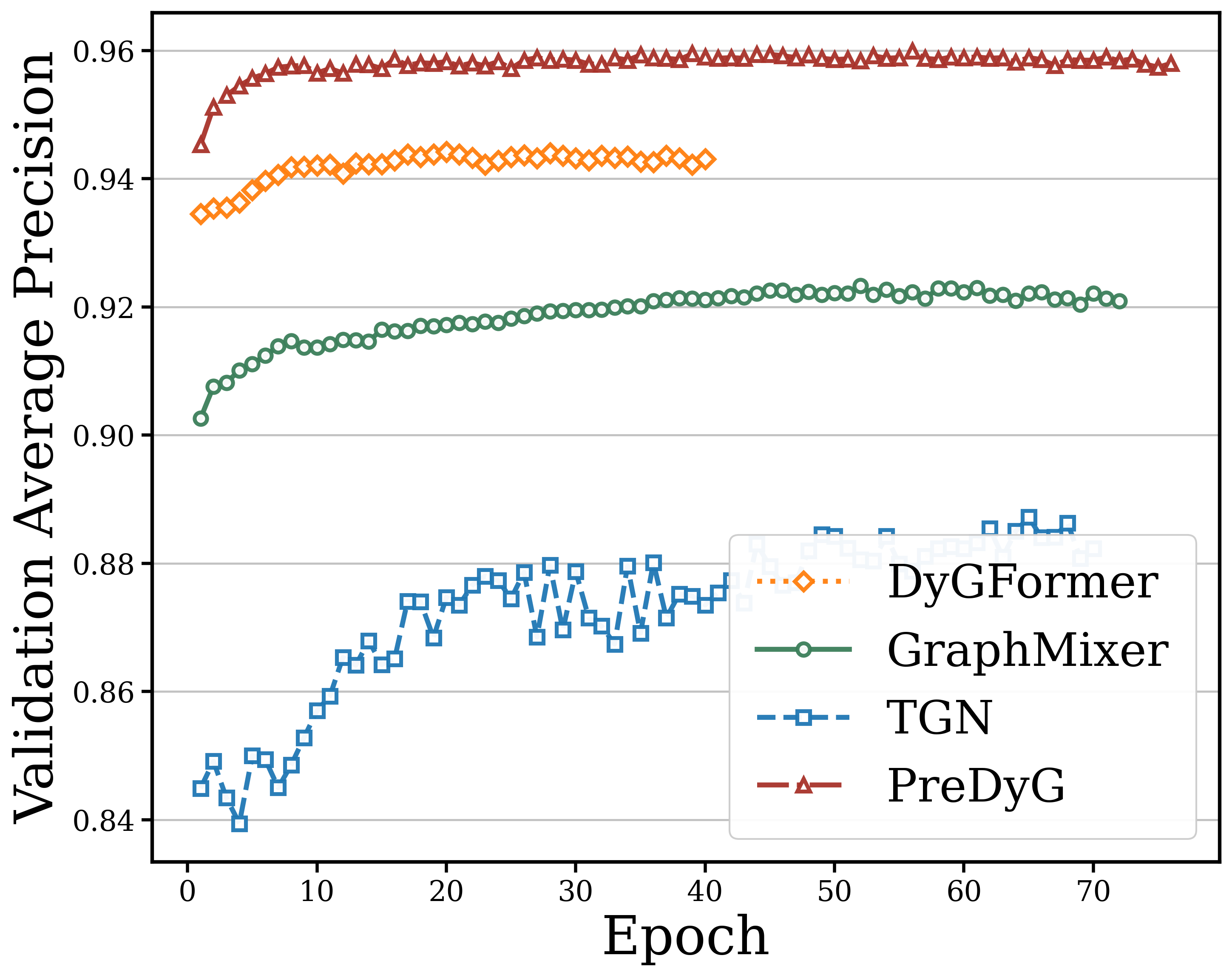}
	\includegraphics[width=0.27\linewidth]{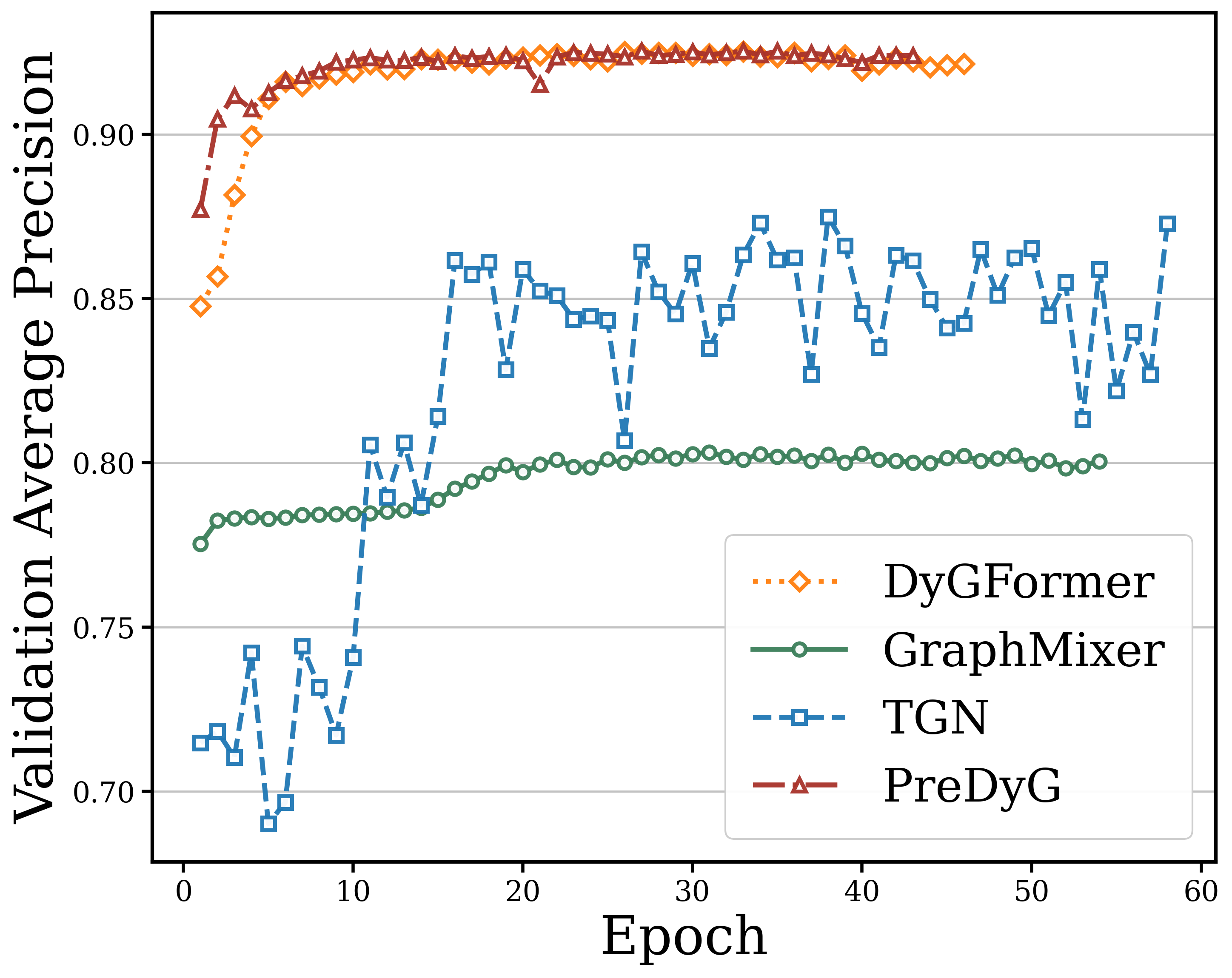}
		\caption{The AP metrics and generalization gap on multiple datasets with the early stopping mechanism were compared: (a) Wikipedia; (b) UCI; (c) Enron.}
	\label{fig_validation_plots}
\end{figure*}

To evaluate whether the compact capacity regime also yields stable optimization and practical savings, we first compare validation trajectories across four interaction domains. As shown in Figure~\ref{fig_validation_plots}, PreDyG reaches a strong plateau rapidly on Wikipedia, and it stabilizes within the first few epochs on UCI and Enron. The behavior is consistent across datasets with different AP ranges and convergence horizons. Compactness therefore does not require unusually slow or unstable optimization. More comparison results are shown in Appendix~\ref{app:AdditionGeneralization}.

\begin{figure}[!htbp]
\centering
\includegraphics[width=\textwidth]{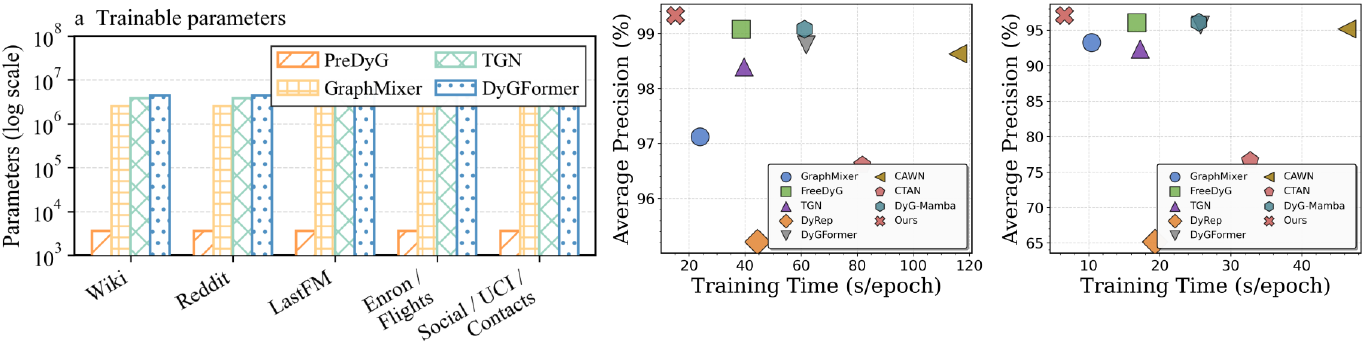}
\caption{Training efficiency. (a) Trainable parameter counts, with baselines grouped into five comparable scales. Validation accuracy versus time per epoch on (b) Wikipedia and (c) UCI.}
\label{fig:efficiency-overview}
\end{figure}

We further compare trainable parameters and measured time per epoch under the common benchmark implementation. As shown in Figure~\ref{fig:efficiency-overview}~(a), GraphMixer, TGN, and DyGFormer use on average \(716\times\), \(1{,}074\times\), and \(1{,}314\times\) as many trainable parameters as PreDyG, respectively. across dataset-specific feature configurations. Figure~\ref{fig:efficiency-overview}~~(b) and~(c) show the running time per epoch of the PreDyG algorithm compared with other baselines. Due to space limitations, we only present a portion of the results. As shown in the figure, we found that in the Wikipedia and UCI datasets, the running time of our algorithm PreDyG was significantly shorter than that of other algorithms, achieving higher validation AP with shorter per-round running time. These results show a favorable accuracy--efficiency trade-off for the evaluated implementation. More results and details can be found in Appendix~\ref{app:AdditionSimplicity}.

\subsubsection{Cross-domain prediction performance}

To test whether our proposed model can maintain excellent prediction quality across a wide range of real-world scenarios, we evaluated it on real-world datasets across different interaction domains with a very small number of parameters and low runtime costs. As shown in Table~\ref{tab:transductive}, our algorithm, PreDyG, achieved state-of-the-art performance on datasets from eight different domains. Furthermore, more complete results regarding the inductive approach can be found in Appendix~\ref{app:InfluenceofdifferentNSS}, where our algorithm also performed exceptionally well in predicting new nodes, achieving the best results. The superior performance in both settings demonstrates that low dimensionality is not limited to a single interaction domain. This is because different domains change the specific content of the predicted state, and the inherent dimensionality required for predicting the state does indeed exist across a broad range of domains.
\begin{table}[!htbp]
\caption{Transductive average precision (\%) on eight domains, reported as mean $\pm$ standard deviation. Best and second-best entries are bold and underlined, respectively.}
\label{tab:transductive}
\centering
\footnotesize
\setlength{\tabcolsep}{2.4pt}
\renewcommand{\arraystretch}{1.10}
\resizebox{\linewidth}{!}{\begin{tabular}{@{}lcccccccc@{}}
\toprule
Method & Wiki & Reddit & LastFM & Enron & SocialEvo & UCI & Flights & Contacts  \\
\midrule
JODIE & $96.50{\pm}.18$ & $98.31{\pm}.12$ & $70.85{\pm}2.11$ & $84.77{\pm}.28$ & $89.89{\pm}.52$ & $89.43{\pm}1.06$ & $95.60{\pm}1.72$ & $95.31{\pm}1.36$  \\
DyRep & $94.86{\pm}.04$ & $98.22{\pm}.07$ & $71.92{\pm}2.20$ & $82.38{\pm}3.32$ & $88.87{\pm}.31$ & $65.14{\pm}2.33$ & $95.29{\pm}.69$ & $95.98{\pm}.15$  \\
TGN & $98.45{\pm}.05$ & $98.63{\pm}.05$ & $77.07{\pm}3.99$ & $86.53{\pm}1.14$ & $93.57{\pm}.21$ & $92.34{\pm}1.02$ & $97.95{\pm}.16$ & $96.89{\pm}.54$  \\
CAWN & $98.76{\pm}.03$ & $99.11{\pm}.01$ & $86.99{\pm}.07$ & $89.56{\pm}.13$ & $84.96{\pm}.09$ & $95.18{\pm}.05$ & $\underline{98.51{\pm}.01}$ & $90.26{\pm}.30$  \\
GraphMixer & $97.25{\pm}.03$ & $97.31{\pm}.01$ & $75.61{\pm}.22$ & $82.25{\pm}.16$ & $93.37{\pm}.07$ & $93.25{\pm}.54$ & $90.99{\pm}.08$ & $91.92{\pm}.03$  \\
DyGFormer & $98.82{\pm}.02$ & $99.11{\pm}.01$ & $93.00{\pm}.12$ & $92.47{\pm}.15$ & $94.47{\pm}.01$ & $95.76{\pm}.16$ & $98.25{\pm}.01$ & $\underline{98.29{\pm}.01}$ \\
FreeDyG & $\underline{99.23{\pm}.01}$ & $99.21{\pm}.01$ & $90.74{\pm}.14$ & $92.51{\pm}.05$ & $\underline{94.54{\pm}.01}$ & $96.03{\pm}.02$ & $98.20{\pm}.01$ & $98.00{\pm}.01$ \\
CTAN & $96.61{\pm}.79$ & $97.21{\pm}.84$ & $86.44{\pm}.80$ & $92.52{\pm}1.20$ & $94.32{\pm}.01$ & $76.64{\pm}4.11$ & $97.24{\pm}.01$ & $96.29{\pm}.03$ \\
DyG-Mamba & $99.08{\pm}.09$ & $\underline{99.12{\pm}.00}$ & $\underline{93.57{\pm}.01}$ & $\underline{92.34{\pm}.12}$ & $94.50{\pm}.01$ & $\underline{96.14{\pm}.14}$ & $98.03{\pm}.02$ & $98.25{\pm}.01$ \\
\midrule
\textbf{PreDyG} & $\bm{99.30{\pm}.01}$ & $\bm{99.34{\pm}.01}$ & $\bm{93.97{\pm}.06}$ & $\bm{93.18{\pm}.07}$ & $\bm{94.68{\pm}.01}$ & $\bm{97.01{\pm}.03}$ & $\bm{98.79{\pm}.01}$ & $\bm{98.32{\pm}.01}$ \\
\bottomrule
\end{tabular}
}
\end{table}

To test the performance of PreDyG on much larger time-graph datasets, we further conducted experiments using TGB. The results are shown in Table~\ref{tab:tgb}, where PreDyG achieved the highest MRR on four large time-graph datasets, including tgbl-review and the tgbl-coin data stream with millions of nodes. This demonstrates that in more complex and longer-term temporal interactions, there still exists an intrinsic dimension required for predicting states, and it plays an important role.
\begin{table}[!htbp]
\caption{MRR (\%) on large TGB datasets. Dashes denote unavailable reported results. Best and second-best entries are bold and underlined, respectively.}
\label{tab:tgb}
\centering
\small
\setlength{\tabcolsep}{5.2pt}
\begin{tabular}{lcccc}
\toprule
Method & tgbl-wiki & tgbl-review & tgbl-coin & tgbl-comment \\
\midrule
CAWN      & 73.04$\pm$.60 & 19.30$\pm$.10 & -- & -- \\
DyRep     & $51.91{\pm}1.95$ & $\underline{40.06{\pm}.59}$ & $45.20{\pm}4.60$ & $28.90{\pm}3.30$ \\
DyGFormer & $77.36{\pm}.31$ & $22.39{\pm}1.52$ & $\underline{75.17{\pm}.38}$ & $67.03{\pm}.14$ \\
TCL       & $\underline{78.11{\pm}.20}$ & $16.51{\pm}1.85$ & $68.66{\pm}.30$ & $\underline{70.11{\pm}.83}$ \\
TGAT      & 59.94$\pm$1.63 & 19.64$\pm$.23 & 60.92$\pm$.57 & 56.20$\pm$2.11 \\
TGN       & 68.93$\pm$.53 & 37.48$\pm$.23 & 58.60$\pm$3.70 & 37.90$\pm$2.00 \\
\midrule
\textbf{PreDyG} & $\bm{78.79{\pm}.38}$ & $\bm{40.30{\pm}.46}$ & $\bm{76.21{\pm}.33}$ & $\bm{71.21{\pm}.72}$ \\
\bottomrule
\end{tabular}
\end{table}
\begin{wraptable}[15]{R}{0.50\textwidth}
	\vspace{-0.7\baselineskip}
	\centering
	\caption{The results of Component ablation experiments.}
	\label{tab:ablation}
	\footnotesize
	\setlength{\tabcolsep}{2.8pt}
	\renewcommand{\arraystretch}{1.08}
	\begin{tabular}{lcccc}
		\toprule
		Variant & Wiki & Reddit & Enron & UCI \\
		\midrule
		Full PreDyG & \textbf{99.30} & \textbf{99.34} & \textbf{93.18} & \textbf{97.01} \\
		w/o NIF & 97.44 & 97.87 & 84.05 & 93.28 \\
		w/o modalities & 98.16 & 98.73 & 90.99 & 95.19 \\
		w/o fusion & 99.07 & 99.18 & 90.43 & 95.14 \\
		\bottomrule
	\end{tabular}
		\caption{The results of NIF sensitivity experiments.}
	\label{tab:sensitivity}
	\footnotesize
	\setlength{\tabcolsep}{2.8pt}
	\begin{tabular}{lcccc}
		\toprule
		Construction & Wiki & Reddit & Enron & UCI \\
		\midrule
		Default NIF & \textbf{99.30} & \textbf{99.34} & \textbf{93.18} & \textbf{97.01} \\
		Learned enum. & 98.00 & 98.21 & 91.88 & 94.30 \\
		Path-NIF & 99.27 & 99.12 & 91.90 & 96.86 \\
		Degree-NIF & 99.34 & 99.20 & 93.12 & 96.90 \\
		\bottomrule
	\end{tabular}
	\vspace{-0.1\baselineskip}
\end{wraptable}

\subsubsection{Ablation Study}
To identify the essential components of PreDyG and test whether its performance depends on a particular neighborhood-statistics construction, we conduct ablation experiments. As shown in Table~\ref{tab:ablation}, removing NIF causes the largest performance loss on every dataset, including a 9.13-point AP drop on Enron. This indicates that PreDyG relies on explicit, low-cost interaction statistics rather than hidden width alone. Removing the node, edge, and temporal modalities, or replacing the learned fusion module, also consistently reduces AP. Table~\ref{tab:sensitivity} varies the construction of neighborhood interaction features. Path-NIF and Degree-NIF remain close to the default construction, whereas learned enumeration is less stable. Thus, the result does not depend on a single hand-designed enumeration rule, although the default NIF construction gives the best average balance.

%% file: iclr2027_supplementary.tex
\clearpage
\appendix

\section{Implementation and Reproducibility Details}
\label{app:implementation}

\subsection{Predictive Dimension Estimation Details}
\label{app:predictive-dimension-settings}

\paragraph{History representation and prediction target.}
Events are sorted by timestamp, preserving their original order in the case of ties, and each stream is limited to its first 200,000 events. Source and destination identities are grouped separately by frequency using a fixed 512-bin rule. Up to 511 frequent identities receive individual bins and the remaining identities share a tail bin, giving the future-distribution representation an ambient dimension of $D_{\mathrm{amb}}=512$. The prediction target is the destination bin of the current event. Its history representation consists of the current source bin and the destination bin of the most recent event involving that source; a separate symbol denotes missing history. The history representation is formed before the current interaction updates the event record.

\paragraph{Conditional distributions and low-rank approximation.}
Conditional probabilities are estimated from history--target counts in the fitting interval using additive smoothing with coefficient $0.5$. History categories absent from this interval are assigned the smoothed destination marginal when constructing the conditional-probability table. We subtract this marginal from the conditional distributions, weight each row by the square root of its empirical history probability, and apply singular value decomposition to the resulting predictive residual matrix. For total dimension $d$, reconstruction retains $d-1$ residual directions and adds back the marginal; $d=1$ therefore represents prediction without history. Reconstructed probabilities are clipped below at $10^{-8}$ and normalized within each row. The history-probability floor used in weighting and inverse weighting is $10^{-12}$.

\paragraph{Risk and dimension selection.}
We evaluate each dimension using the mean Brier risk,
\begin{equation}
R(d)=\frac{1}{n}\sum_{k=1}^{n}\sum_{f=1}^{K}
\left[q_d(f\mid H_k)-\mathbf{1}\{F_k=f\}\right]^2,
\label{eq:app-brier-risk}
\end{equation}
where $n$ is the number of evaluation events, $K$ is the number of destination categories, $H_k$ is the history representation before event $k$, $F_k$ is its destination category, $q_d$ is the predictive distribution at dimension $d$, and $\mathbf{1}\{\cdot\}$ denotes the indicator function. Within a candidate set, the selected dimension is
\begin{equation}
d^\star=\min\left\{d\in\mathcal D:
R(d)-R_{\min}\leq 0.05\left[R(1)-R_{\min}\right]\right\},
\qquad R_{\min}=\min_{d\in\mathcal D}R(d),
\label{eq:app-dimension-selection}
\end{equation}
where $\mathcal D$ is the candidate dimension set, $R(1)$ is the no-history risk, $R_{\min}$ is the lowest risk attained within that set, and $d^\star$ is the smallest dimension retaining at least $95\%$ of the attainable risk reduction. If this reduction is at most $10^{-9}$, the selected dimension is set to 1.

The first $40\%$ of events fit the past--future operator and the next $20\%$ select $\widehat d_{\mathrm{pred}}$ on the grid $\{1,2,4,8,16,32,64,128,256,512\}$. We then refit the operator on the first $70\%$ of events and apply the same grid and $5\%$ risk criterion to the final $30\%$, producing the independent held-out estimate $d_{\mathrm{hold}}$. Both estimates therefore use the same prediction target, ambient representation, rank grid, and tolerance, but are selected on disjoint chronological evaluation intervals.

\subsection{Scale and Compression Statistics}
\label{app:scale-compression-settings}

\paragraph{Scale statistics.}
All competing statistics are computed on the same event prefix used for predictive-dimension estimation. Event count is the number of interactions in this prefix; source and destination counts enumerate distinct identities on their respective sides. Distinct-pair count enumerates directed source--destination pairs, counting repeated interactions only once. The node-scale correlation reported by the implementation uses the number of distinct source identities.

\paragraph{Event-stream encoding.}
Source and destination identities are relabeled separately in order of first occurrence. Each event is serialized as its source identifier, destination identifier, and time since the preceding event, with comma-separated fields and one event per line in a UTF-8 byte stream. Time differences are measured in the dataset's original timestamp units, clipped below at zero, and rounded down to integers; the first event uses time zero as its reference.

The same byte stream is compressed with zlib and bz2 at compression level 9 and with lzma at preset 9. The resulting statistic is
\begin{equation}
\ell_c=\frac{8\,|c(S)|}{N},
\label{eq:app-compression-length}
\end{equation}
where $S$ is the serialized event stream, $c$ is the specified compressor, $|c(S)|$ is the compressed size in bytes, $N$ is the number of events, and $\ell_c$ is the description length in bits per event. We compute the Spearman correlation of each statistic with $d_{\mathrm{hold}}$ across the eight datasets used in the capacity analysis. Tied values receive average ranks.

\subsection{Comparison with Scale and Compression Statistics}
\label{app:statistic-comparison}

We compare predictive dimension, event count, source-node count, distinct-pair count, and event-stream description lengths by their Spearman correlations with the independently estimated held-out dimension $d_{\mathrm{hold}}$. Statistic-computation details are provided in Appendix~\ref{app:scale-compression-settings}.

Figure~\ref{fig:statistic-comparison} reports the resulting rank associations. $\widehat d_{\mathrm{pred}}$ has the strongest association with $d_{\mathrm{hold}}$ ($\rho=0.892$). The corresponding correlations are $0.546$ for source-node count, $0.464$ for distinct-pair count, and $0.188$ for event count. The zlib, bz2, and lzma description lengths each yield $\rho=0.082$.

\begin{figure}[!htbp]
	\centering
	\includegraphics[width=0.55\textwidth]{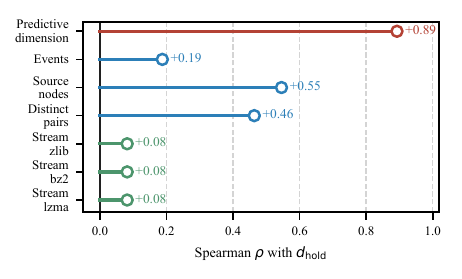}
	\caption{\textbf{Competing data statistics.} Spearman correlation of each statistic with $d_{\mathrm{hold}}$ across the eight datasets used in the capacity analysis.}
	\label{fig:statistic-comparison}
\end{figure}

\section{Additional Details for Predictive-Dimension Analysis}
\label{app:nulls}

\subsection{Dependency Interventions}

The \emph{time-shuffle} null permutes timestamps and stably resorts the stream; \emph{event-order shuffle} permutes rows directly; \emph{destination shuffle} and \emph{source shuffle} independently permute the respective endpoint column; and \emph{edge-pair shuffle} permutes intact endpoint pairs across event positions. These interventions preserve event count, and pair shuffling additionally preserves the multiset of edges. They are not interchangeable: each removes a different carrier of dependence.

\subsection{Compression Controls}

The event-stream serialization and compression settings are specified in Appendix~\ref{app:scale-compression-settings}. Additional compression controls use a static encoding of sorted unique source--destination pairs and a residual encoding of whether the destination repeats the source's previous destination together with the clipped inter-event time.

\section{Addition Simplicity Experiments}
\label{app:AdditionSimplicity}
\begin{figure*}
	\centering
	\includegraphics[width=0.4\linewidth]{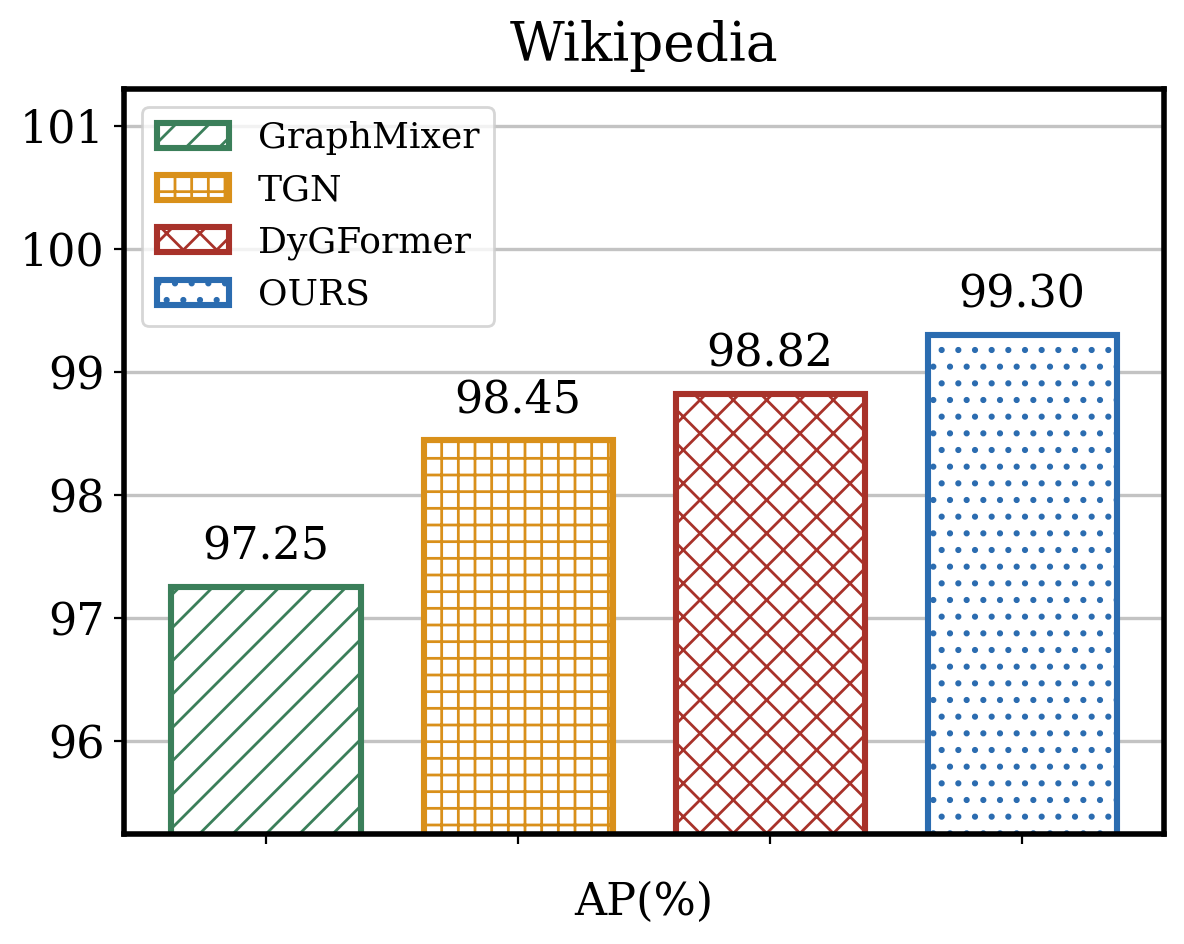}
	\includegraphics[width=0.4\linewidth]{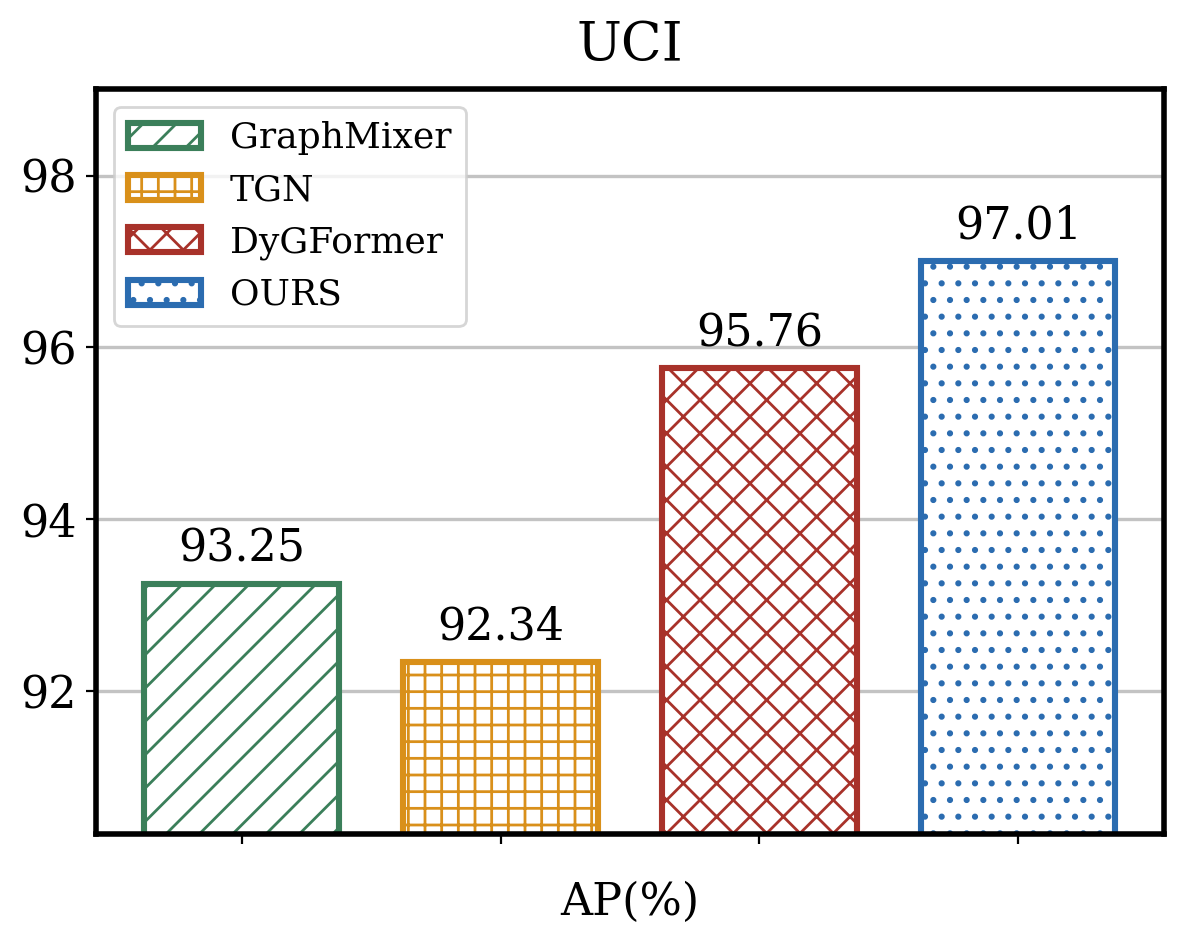}
	\includegraphics[width=0.4\linewidth]{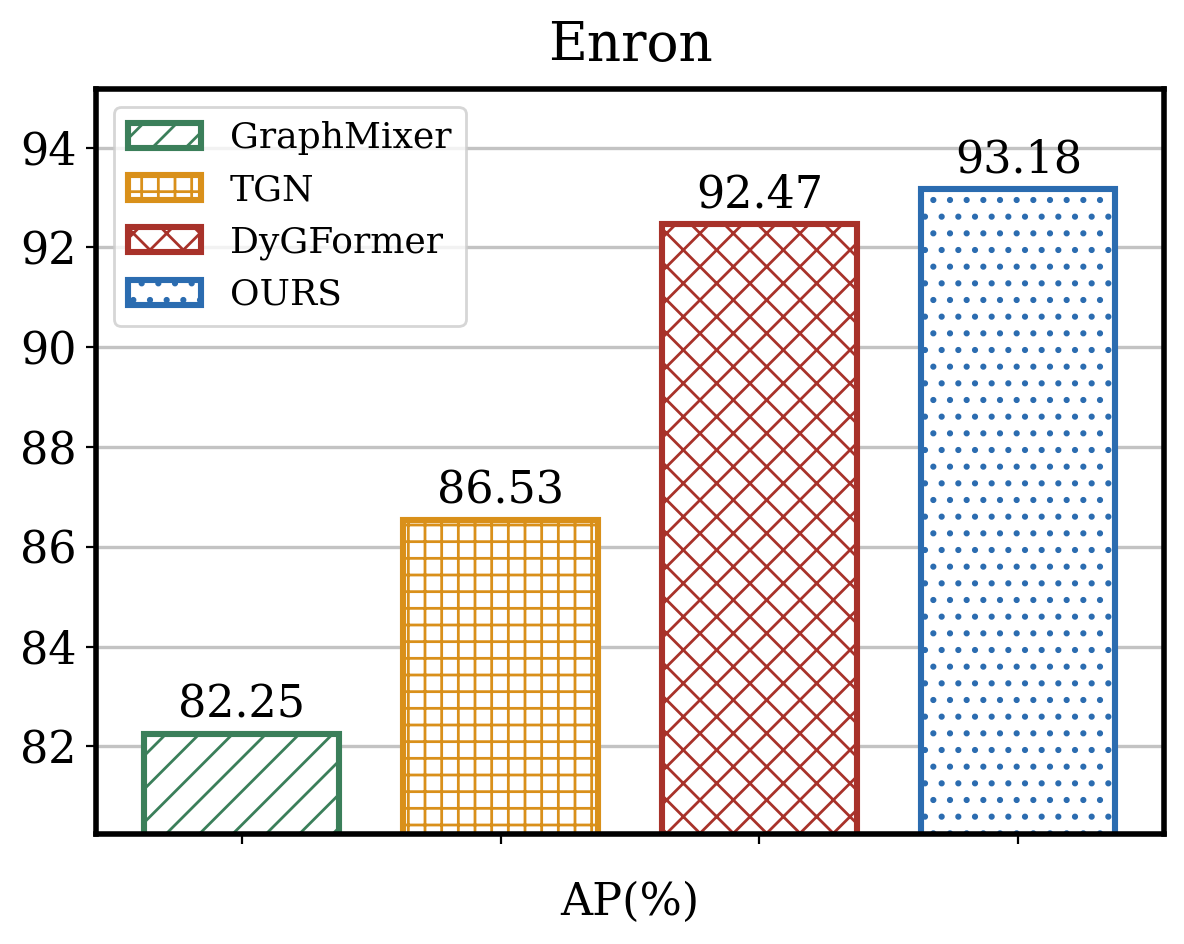}
	\includegraphics[width=0.4\linewidth]{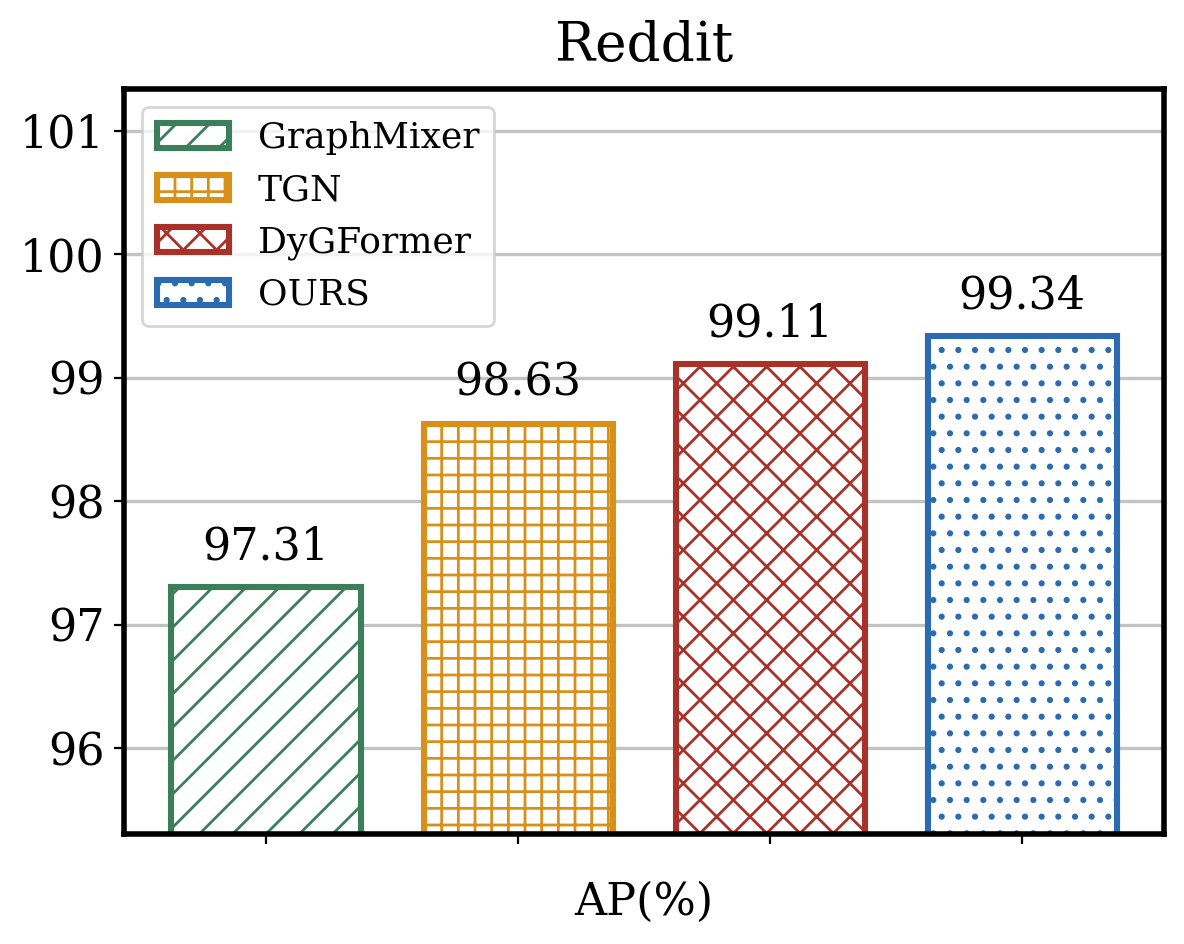}
	\caption{Compared to the performance of lightweight baselines, our method achieves state-of-the-art results on all datasets.}
	\label{fig_Simplicity}
\end{figure*}

\begin{figure*}
	\centering
	\includegraphics[width=0.4\linewidth]{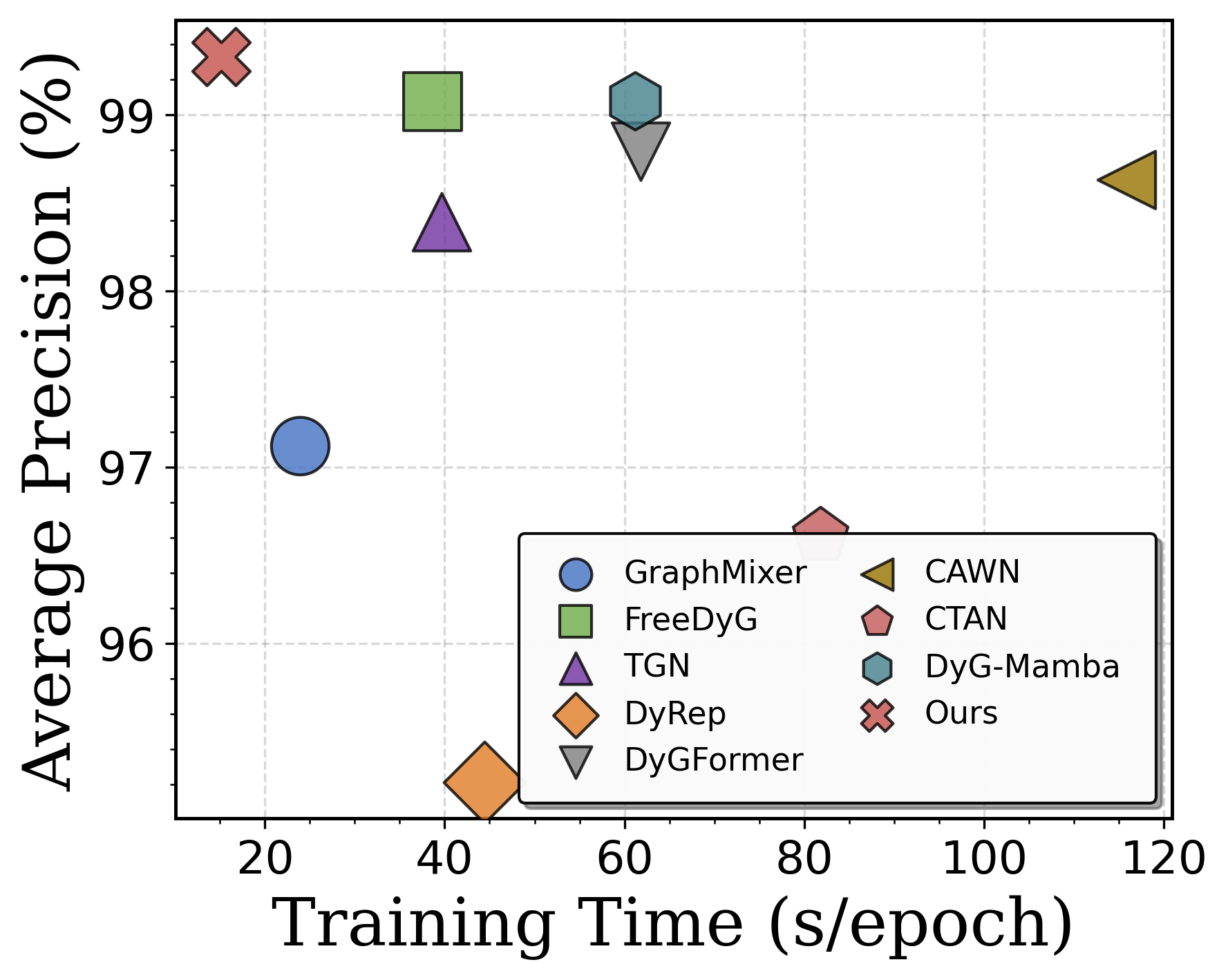}
	\includegraphics[width=0.4\linewidth]{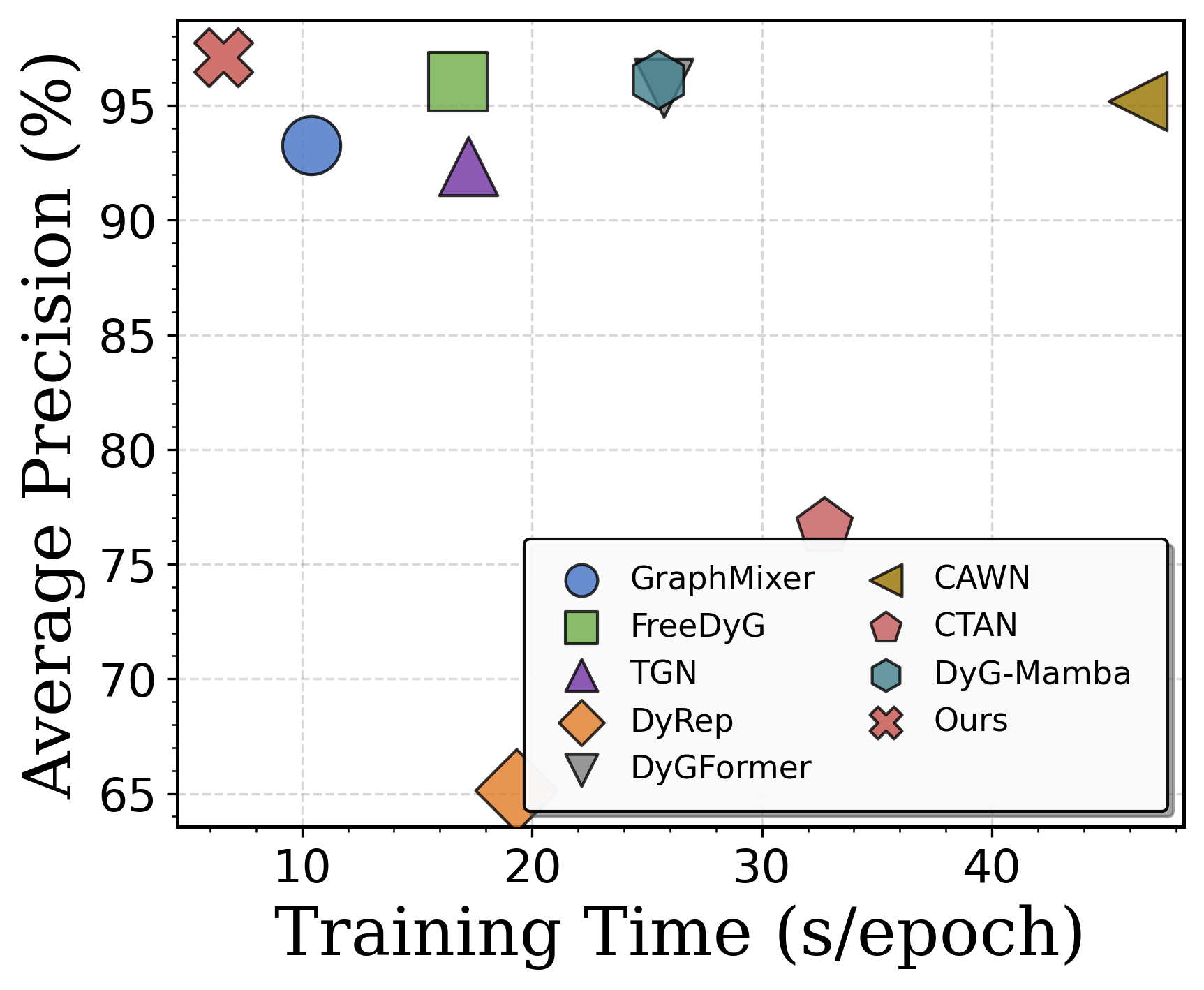}
	\includegraphics[width=0.4\linewidth]{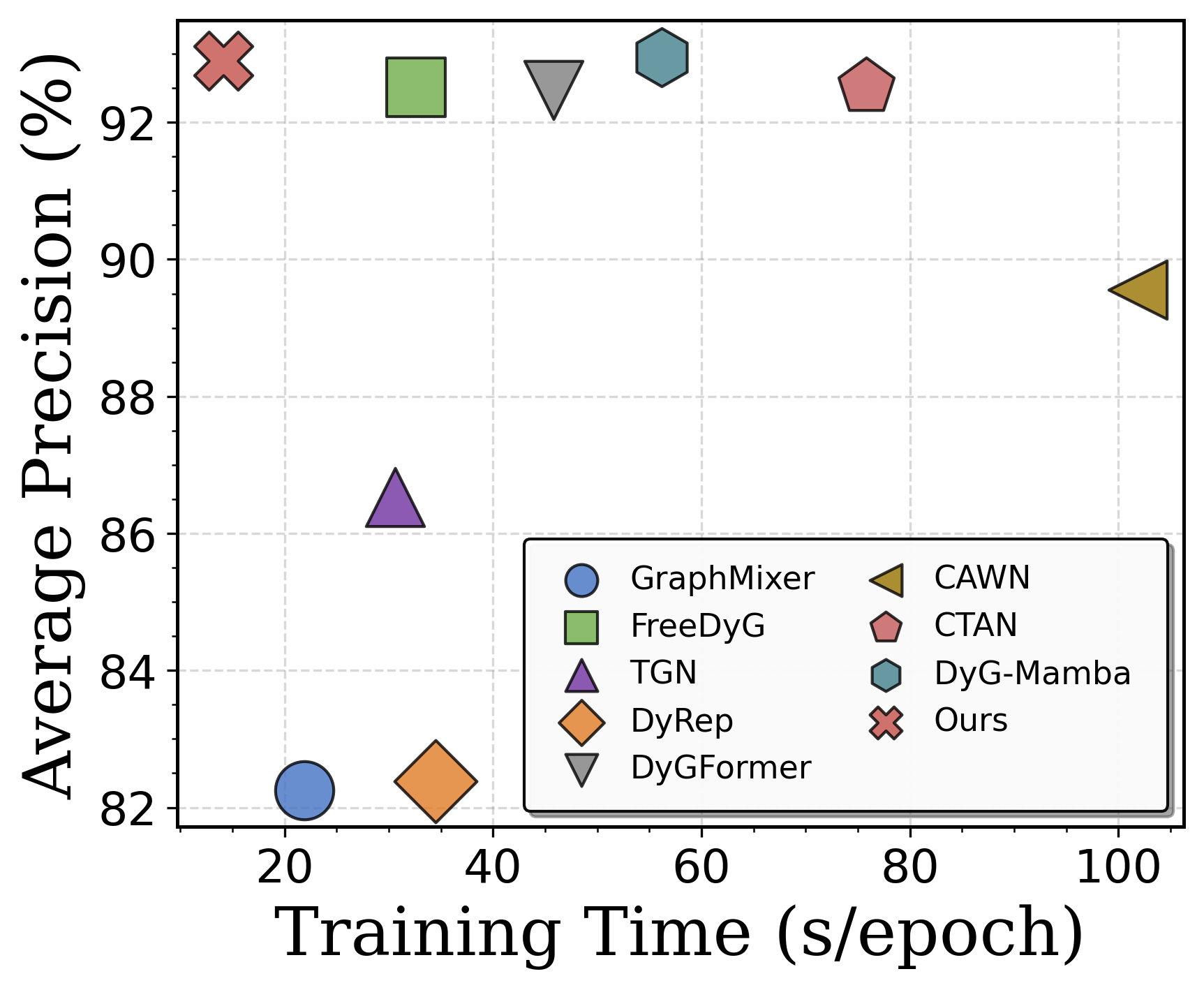}
	\includegraphics[width=0.4\linewidth]{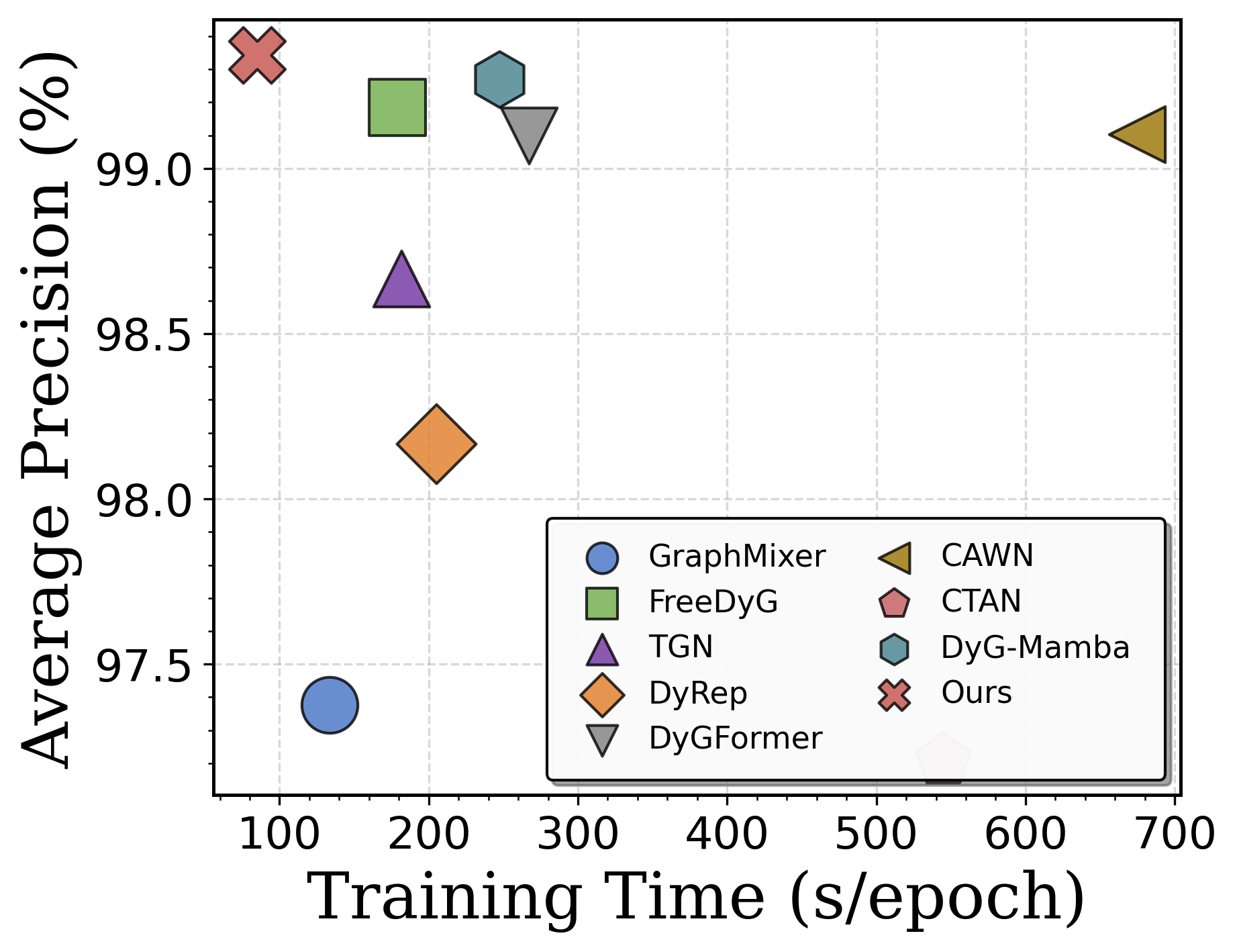}
	\caption{Training time comparison for one epoch across different baselines on multiple datasets: (a) Wikipedia; (b) UCI; (c) Enron; (d) Reddit.}
	\label{fig_Trainingall}
\end{figure*}

We report additional experimental results that are not fully included, as shown in Figure \ref{fig_Simplicity}. The figure compares performance against lightweight baseline methods on four datasets, namely \textit{Wikipedia}, \textit{UCI}, \textit{Enron}, and \textit{Reddit}. The results show that our method achieves state-of-the-art performance on all datasets while maintaining a more concise model architecture than the baselines. Notably, despite its lower complexity, our method consistently attains higher Average Precision than other more complex baseline methods. Figure \ref{fig_Trainingall} provides supplementary results on training-time efficiency and reports the time cost per epoch for our method compared with other baselines. The figure shows that our method not only delivers strong predictive performance but also offers clear advantages in training efficiency, further validating its practicality and scalability.

\section{Experimental Details}
\label{app:ExperimentalDetails}
\input{tab_tg_data}
\subsection{Comprehensive Dataset Collection}
\label{app:ComprehensiveDatasetCollection}
Our model-side evaluation includes eight conventional temporal graph datasets and four Temporal Graph Benchmark (TGB) datasets.

\noindent\textbf{Wikipedia}: This dataset records the collaborative editing activities on the Wikipedia platform. It constructs a bidirectional network where users interact with articles through timestamped editing events. Each interaction includes language features extracted from the edited content as well as temporal user status labels indicating editing permissions. This dataset covers multiple years and reveals the situations in aspects such as knowledge creation, malicious destruction detection, and the evolution of community governance in large-scale collaborative systems. 

\noindent\textbf{Reddit}: This dataset simulates users' participation in online communities through the submission of posts and interaction with comments. This bidirectional temporal network can capture the dynamic participation of the community, including rich time-varying information about content categories, user activity patterns, and user statuses (including management events). This dataset can be used to analyze the dynamics of social media ecosystems and the formation process of communities.

\noindent\textbf{LastFM}: This dataset records the long-term music streaming behavior of users and artists based on their listening events. This two-part temporal network reveals the evolution of preferences, discovery patterns, and consumption habits over time, including metadata such as play counts, music genres, and demographic information, and is highly suitable for research in recommendation systems.

\noindent\textbf{Enron}: This dataset, through email communication networks spanning multiple years, reveals the enterprise communication patterns within large organizations. This dynamic graph dataset records the evolution of organizational hierarchical structures, information flow patterns, and communication dynamics during normal operations and crisis periods, thereby enabling the analysis of the network structure of the enterprise.

\noindent\textbf{Social Evo.}: This dataset captures the intricate spatio-temporal interaction patterns among college students through proximity detection of mobile devices. The time network presents physical proximity events with high temporal resolution, revealing daily behavioral patterns, the formation of social groups, and the impact of academic schedules on social interactions.

\noindent\textbf{Contact}: This dataset is based on the concept of proximity detection, using wearable sensors to monitor the close interactions among student groups. Its accuracy can reach the minute level. The dataset contains detailed metadata about the interaction scenarios and is capable of analyzing social contact patterns, interaction rhythms, and the evolution of community structures.

\noindent\textbf{Flights}: This dataset records the evolution of the aviation network during major global events, covering the adjustments of routes and changes in traffic patterns in the international flight network. Its temporal dynamics reflect the operational strategies of airlines, seasonal patterns, and crisis response measures, providing important insights into the stability of the transportation network.

\noindent\textbf{UCI}: This dataset records the communication behaviors within the university student population through various digital communication channels. The time graph depicts the interactions among students and includes metadata about the types of communication and academic backgrounds, thereby revealing how academic pressure and social events affect communication patterns and the formation of interpersonal relationships.

The four TGB datasets are described below~\cite{tgb2023}.

\noindent\textbf{tgbl-wiki.} This dataset records timestamped editing interactions between Wikipedia users and pages over one month. It forms a bipartite graph, with each event representing a user editing a page. The prediction task is to identify the page a user will interact with at a given time.

\noindent\textbf{tgbl-review.} This dataset contains timestamped reviews of Amazon electronics products. Users and products form the two node sets, and each interaction represents a user reviewing a product. The prediction task is to identify the product a user will review at a given time.

\noindent\textbf{tgbl-coin.} This dataset records cryptocurrency transfers between blockchain addresses. Nodes represent addresses, and directed events represent timestamped transfers. The prediction task is to identify the destination address of a future interaction.

\noindent\textbf{tgbl-comment.} This dataset represents a directed Reddit reply network. Nodes are users, and each timestamped event records one user replying to another. The prediction task is to identify the recipient of a user's reply.

\subsection{Baseline Algorithm Landscape}
\label{app:BaselineAlgorithmLandscape}
\noindent\textbf{DyRep}\cite{ctdg_message_passing}: This algorithm establishes the basic method by combining the recurrent neural network architecture and the temporal attention mechanism. It maintains the constantly changing node representations through continuous state updates, which can capture the changes in the local neighborhood as well as the global temporal patterns. By adopting a dual encoder architecture and combining the attention mechanism, it selectively focuses on relevant historical events when updating the node states.

\noindent\textbf{TGN}\cite{ctdg_memory}: This algorithm combines a trainable memory library and a message-passing framework to achieve temporal graph learning with enhanced memory capabilities. The architecture sets up dedicated storage modules for each node to store compressed historical information, and uses an attention-based aggregation method to combine the received messages with the stored historical background information. 

\noindent\textbf{CAWN}\cite{cawn2021}: Causal Anonymous Walk-based methods extract meaningful temporal pathways that preserve causality constraints while maintaining structural anonymity. The method generates anonymous walk sequences through time-aware sampling procedures, followed by anonymous encoding that creates generalizable structural representations while preserving essential temporal properties.

\noindent\textbf{GraphMixer}\cite{graphmixer2023}: This algorithm represents a shift towards a time-encoding encoder based on simplified multi-layer perceptrons. This encoder can achieve comparable performance to similar methods while also reducing computational complexity. GraphMixer emphasizes a fixed time-encoding strategy and employs specialized mixing layers. These mixing layers process temporal and structural information independently through separate paths, and then combine them through a learnable fusion mechanism.

\noindent\textbf{DyGFormer}\cite{ctdg_transformer}: Dynamic Graph Transformer has advanced the development of converter-based temporal graph learning by adopting a specialized architecture with an adjacent co-occurrence encoding scheme and a stitching technique. This method combines temporal position encoding and multi-head attention mechanisms, which have been optimized for temporal graph data and can efficiently handle extended temporal sequences.

\noindent\textbf{FreeDyG}\cite{freedyg2024}: Frequency-domain Dynamic Graph learning incorporates Fourier transformation integration to conduct frequency-domain analysis on temporal graph patterns. This method converts time series into frequency-domain representations, where periodic patterns become more prominent. Additionally, it employs learnable frequency-domain operations and a frequency-based message-passing mechanism.

\noindent\textbf{CTAN}\cite{ctan2024}: This algorithm overcomes the limitations of using specially designed mechanisms to capture the complex temporal dependencies between long sequences, which are applicable for long sequence modeling. This method can handle dependencies at different granularities while maintaining computational feasibility.

\noindent\textbf{DyG-Mamba}\cite{dygmamba2025}: Dynamic Graph Mamba represents recent developments through state-space model integration, providing efficient mechanisms for processing very long temporal sequences. The method adapts state-space modeling principles with selective state updates based on temporal interactions, achieving linear computational complexity while maintaining competitive performance on temporal graph learning benchmarks.

In addition, we briefly introduce other methods shown in Figure~1, such as EvolveGCN TP-Net and DGM. EvolveGCN evolves GCN parameters over time with an RNN to model dynamic graphs without maintaining explicit node embeddings, enabling adaptation to changing node sets~\cite{evolvegcn2020}. TPNet learns time-decayed temporal-walk encodings with efficient random-feature propagation for fast and accurate temporal link prediction~\cite{tpnet2024}. A cross-geometric dynamic graph framework that jointly leverages Euclidean and hyperbolic embeddings with substructure-aware alignment and temporal aggregation to model evolving node and relation dynamics~\cite{sunanalytical}.
\subsection{Addition Generalization Experiments}
\label{app:AdditionGeneralization}
\begin{figure*}
	\centering
	\includegraphics[width=0.32\linewidth]{validation_plots/wikipedia_validation_average_precision.png}
	\includegraphics[width=0.32\linewidth]{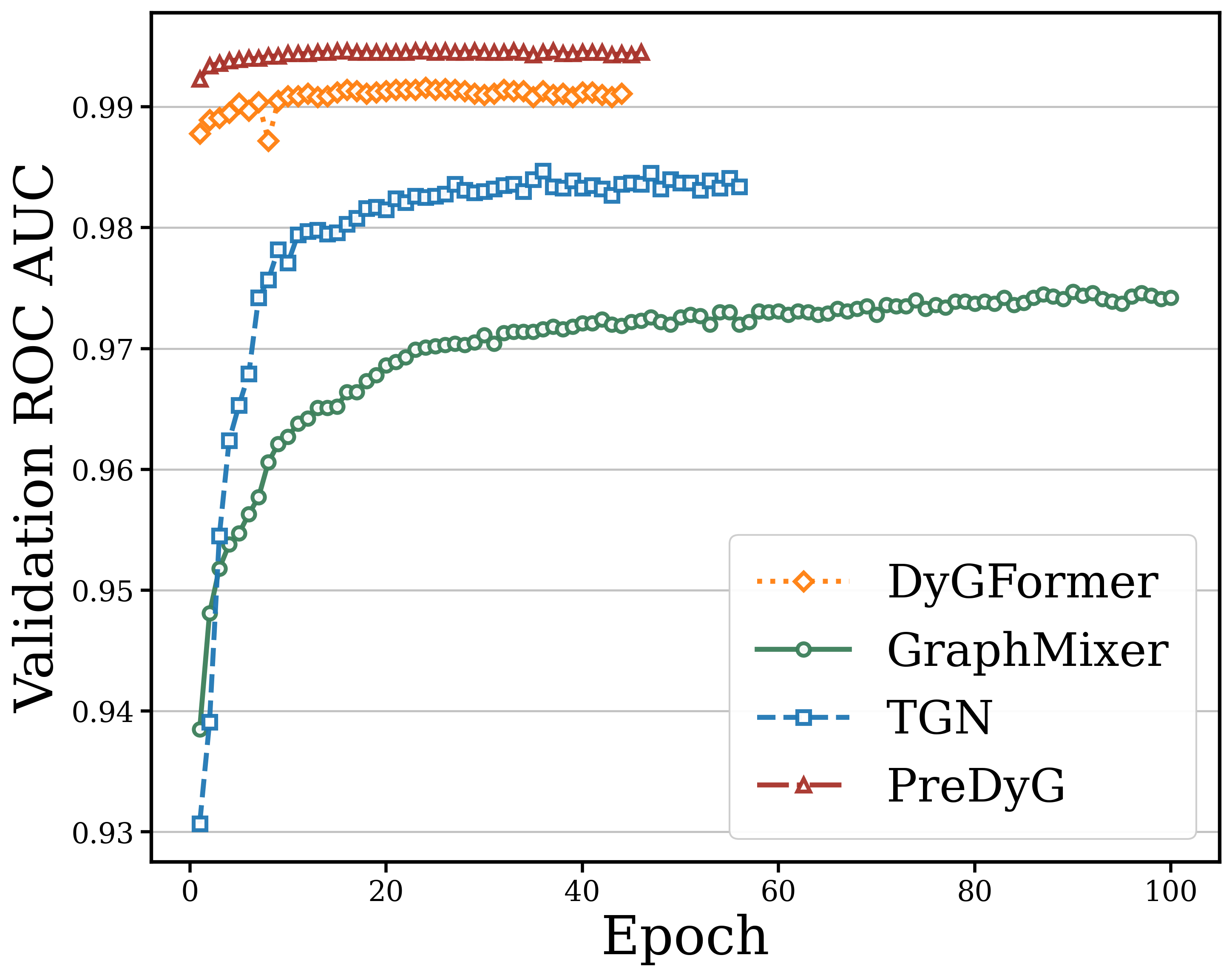}
	\includegraphics[width=0.32\linewidth]{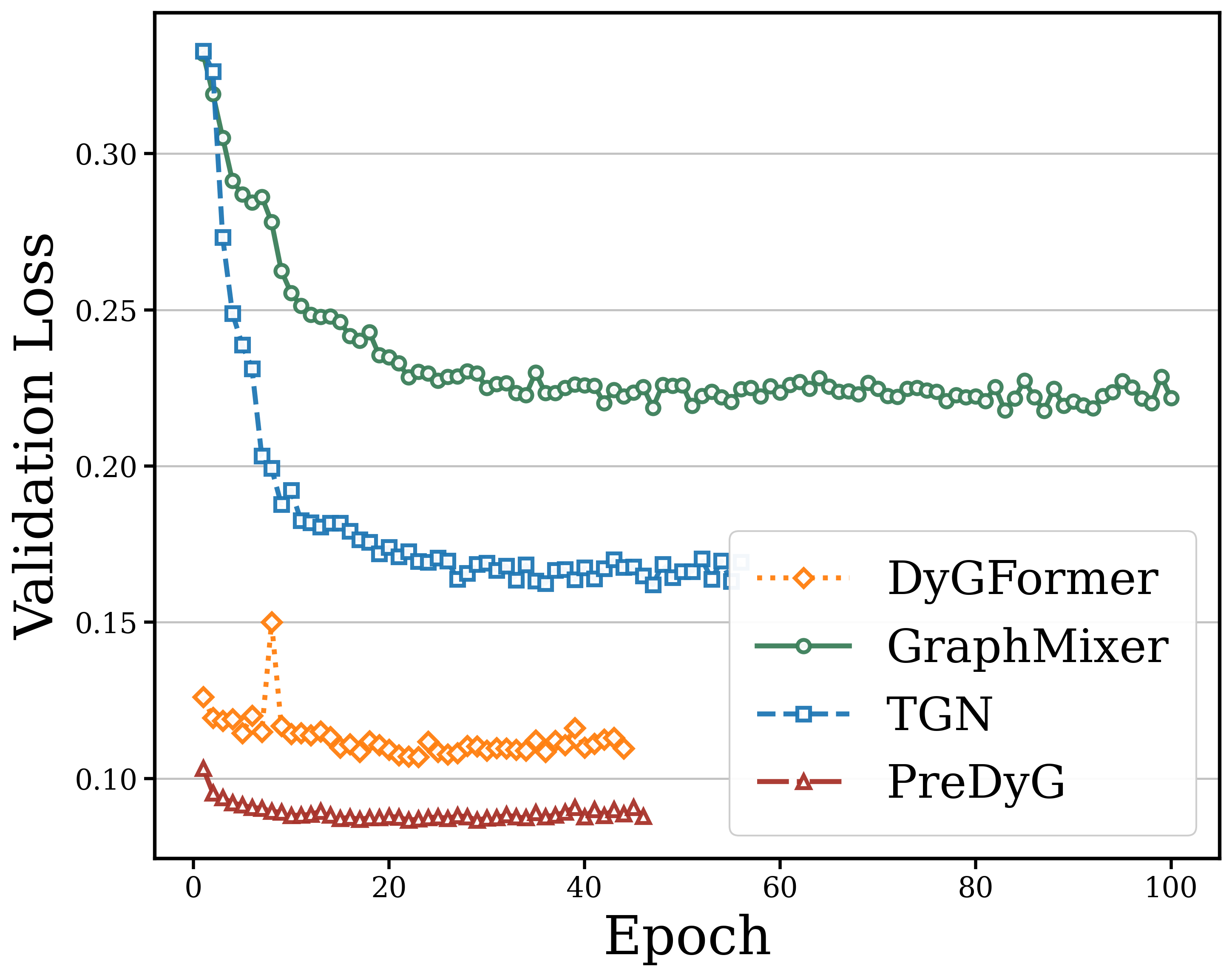}
	
	\includegraphics[width=0.32\linewidth]{validation_plots/uci_validation_average_precision.png}
	\includegraphics[width=0.32\linewidth]{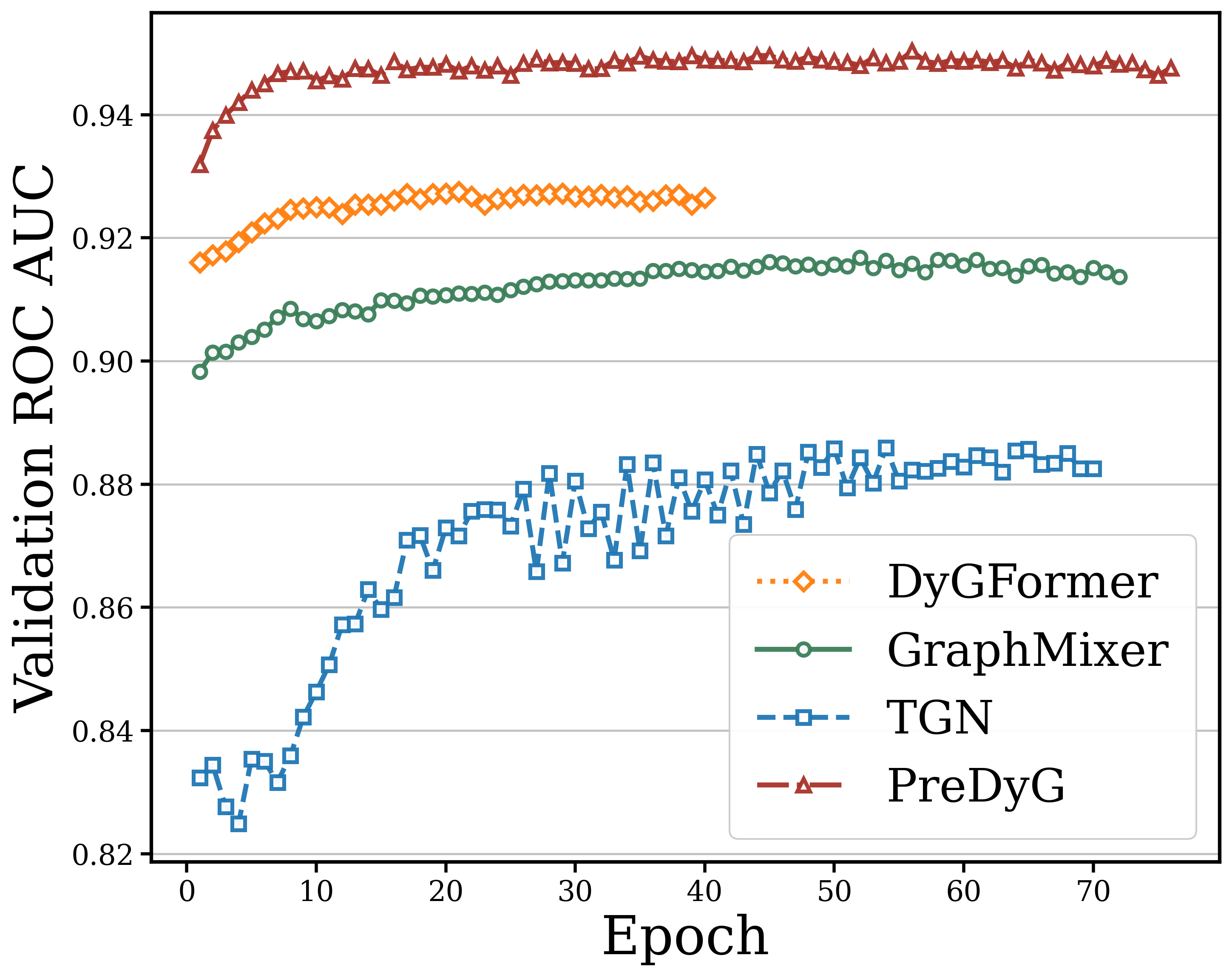}
	\includegraphics[width=0.32\linewidth]{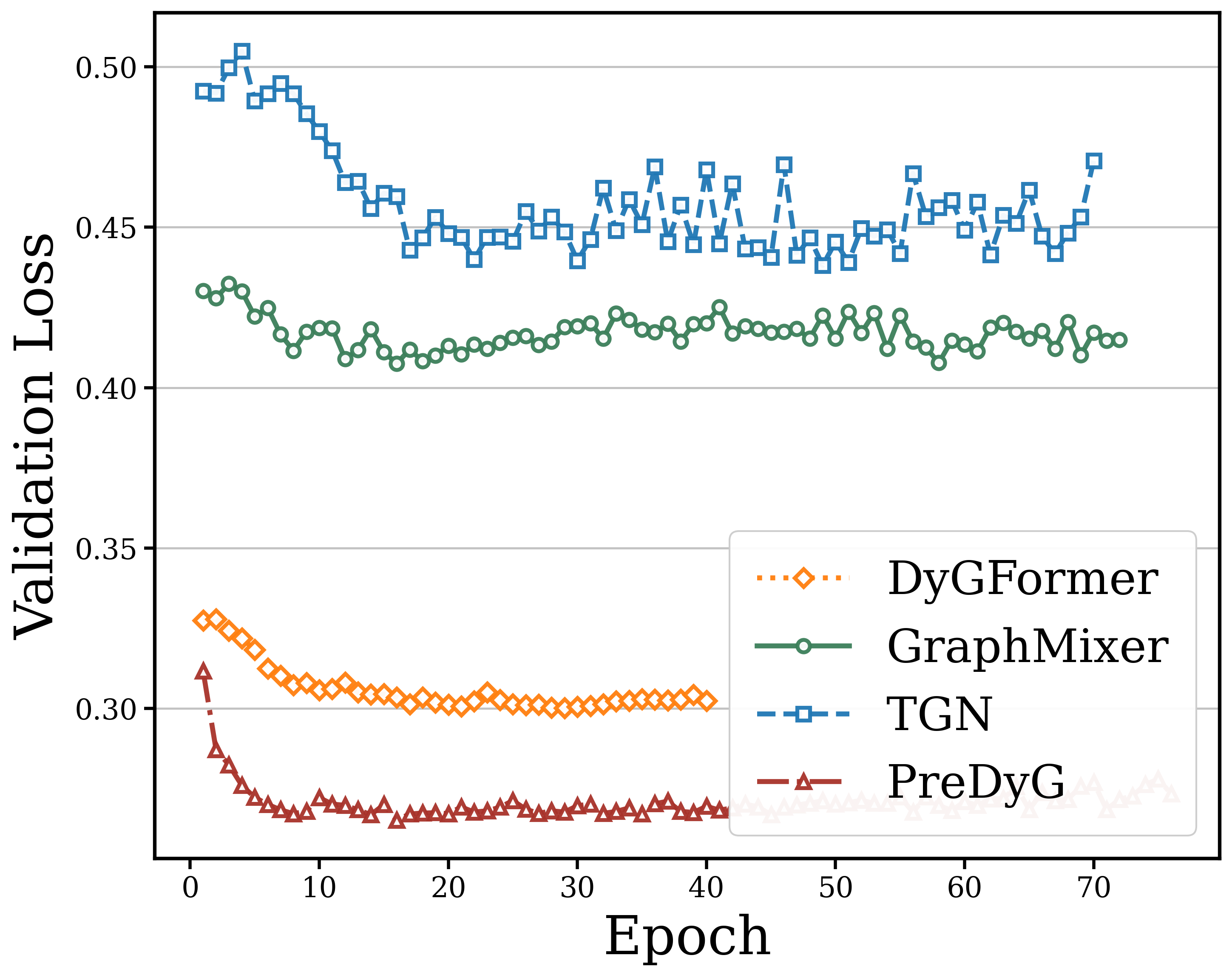}
	
	\includegraphics[width=0.32\linewidth]{validation_plots/enron_validation_average_precision.png}
	\includegraphics[width=0.32\linewidth]{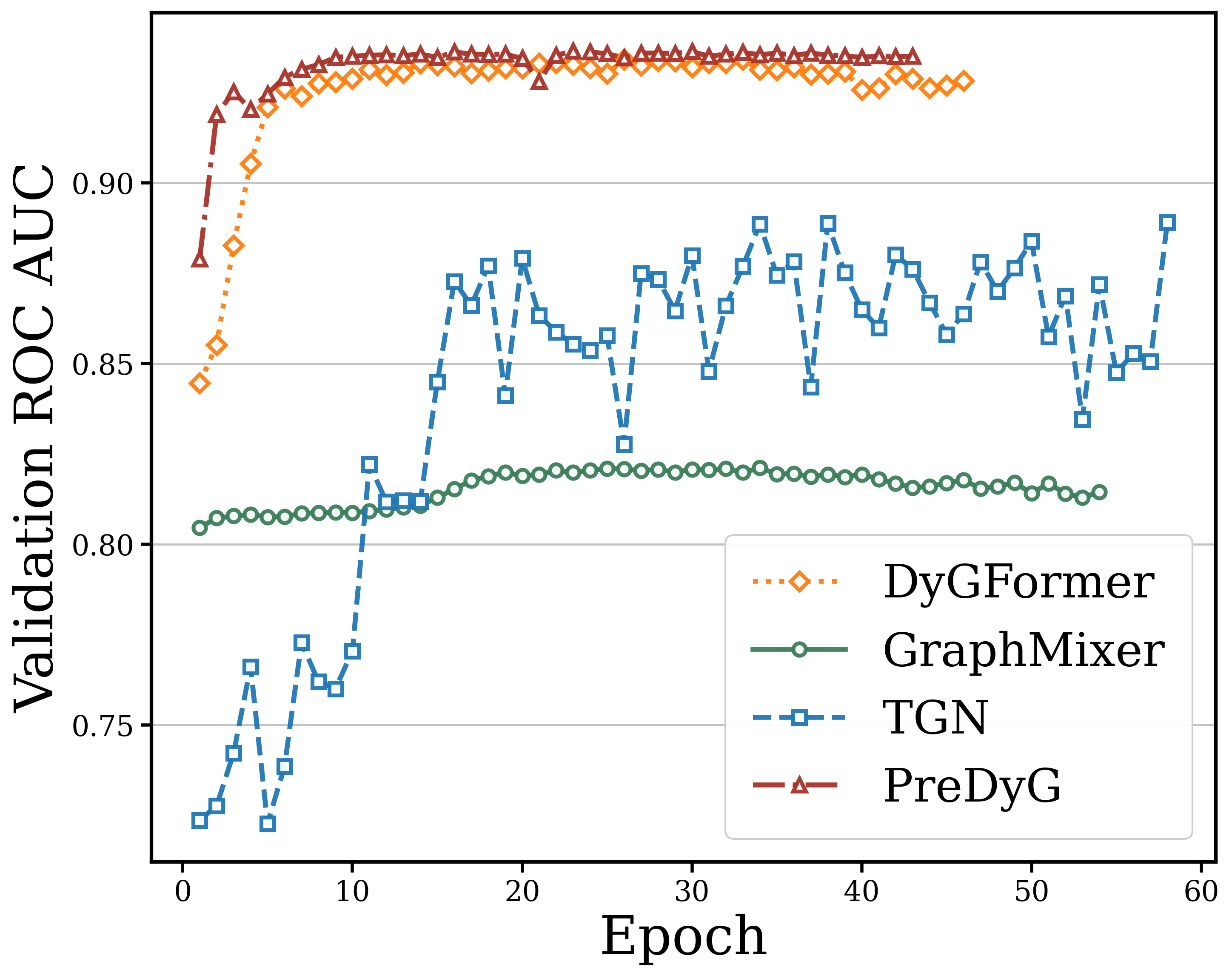}
	\includegraphics[width=0.32\linewidth]{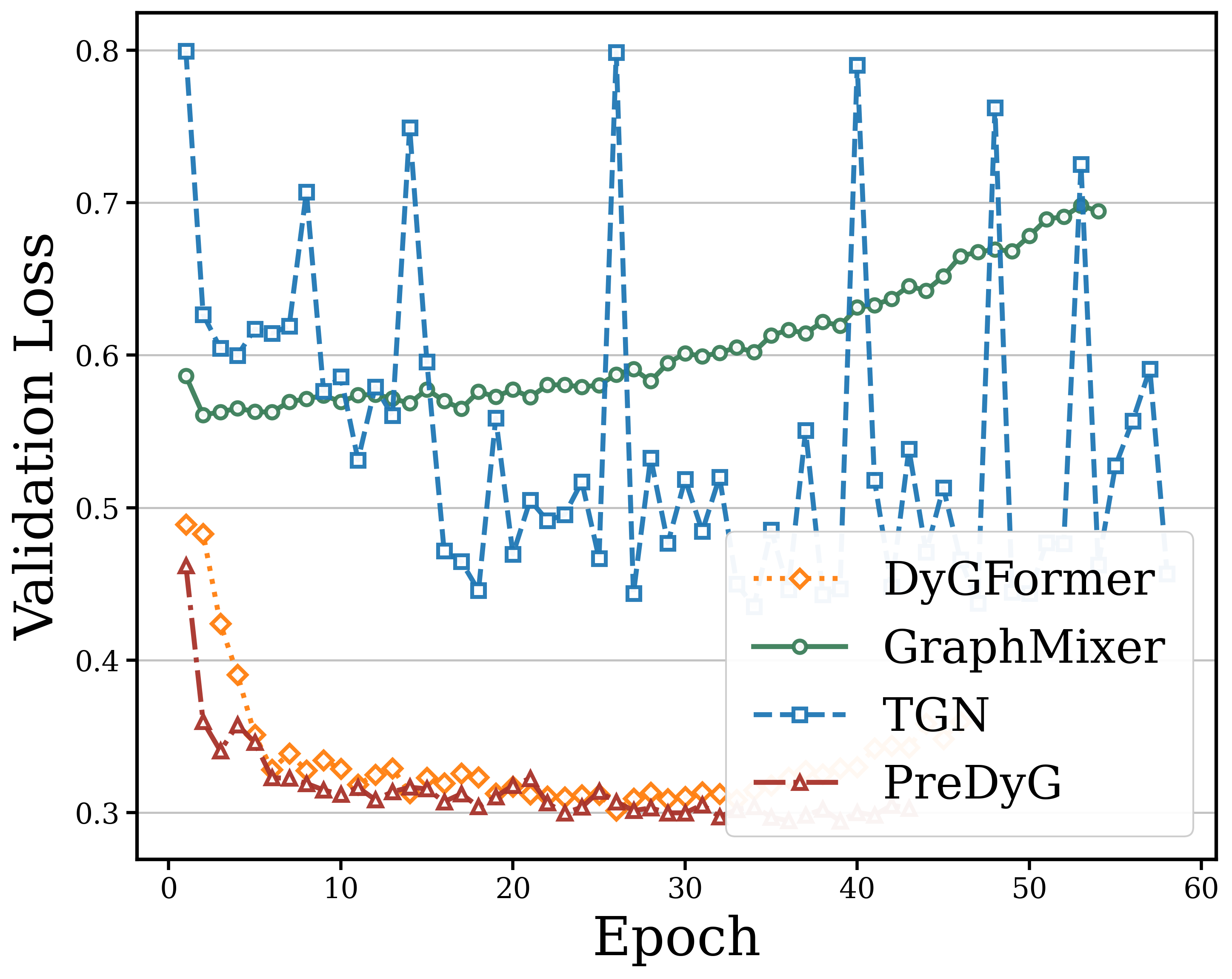}
	
	\includegraphics[width=0.32\linewidth]{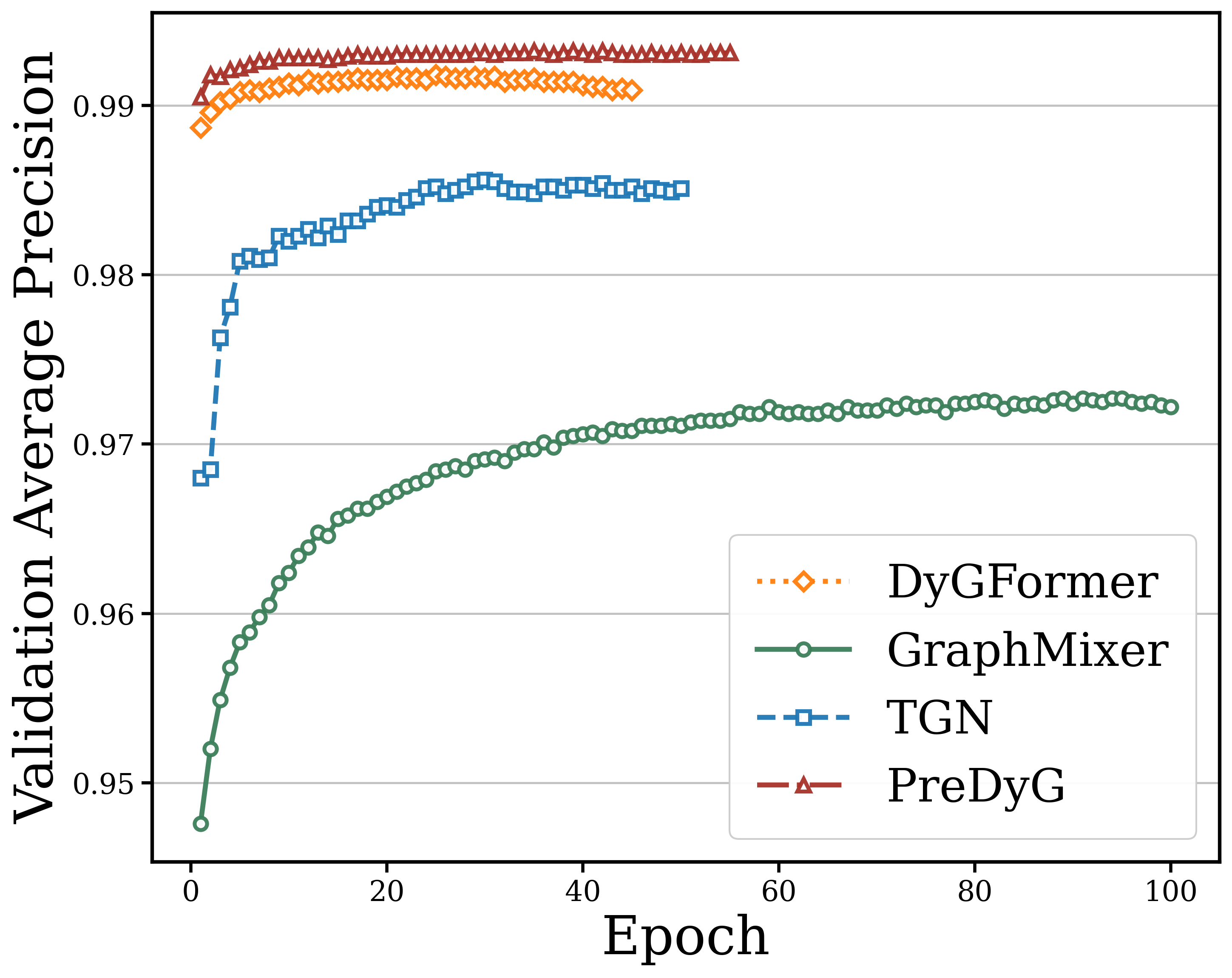}
	\includegraphics[width=0.32\linewidth]{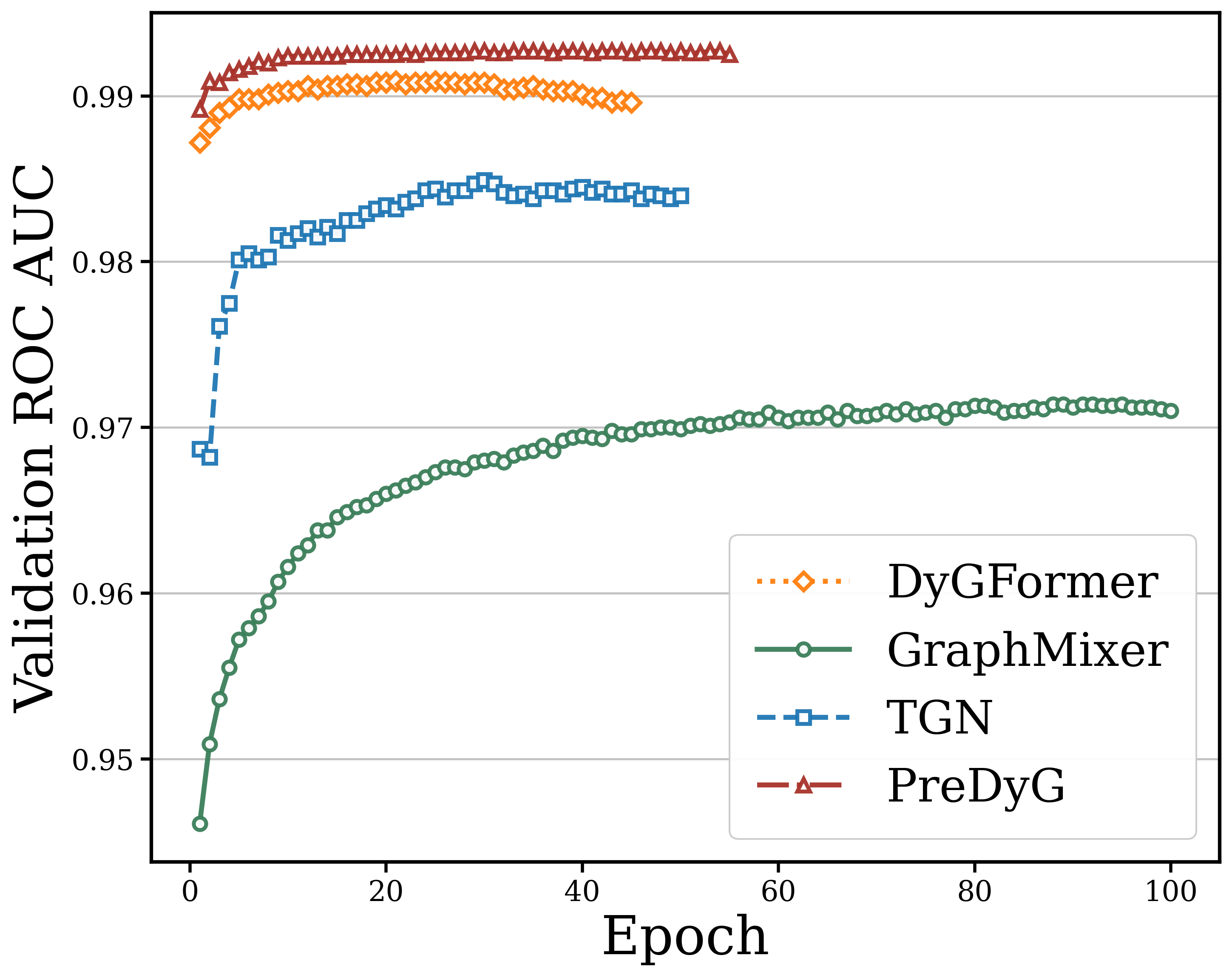}
	\includegraphics[width=0.32\linewidth]{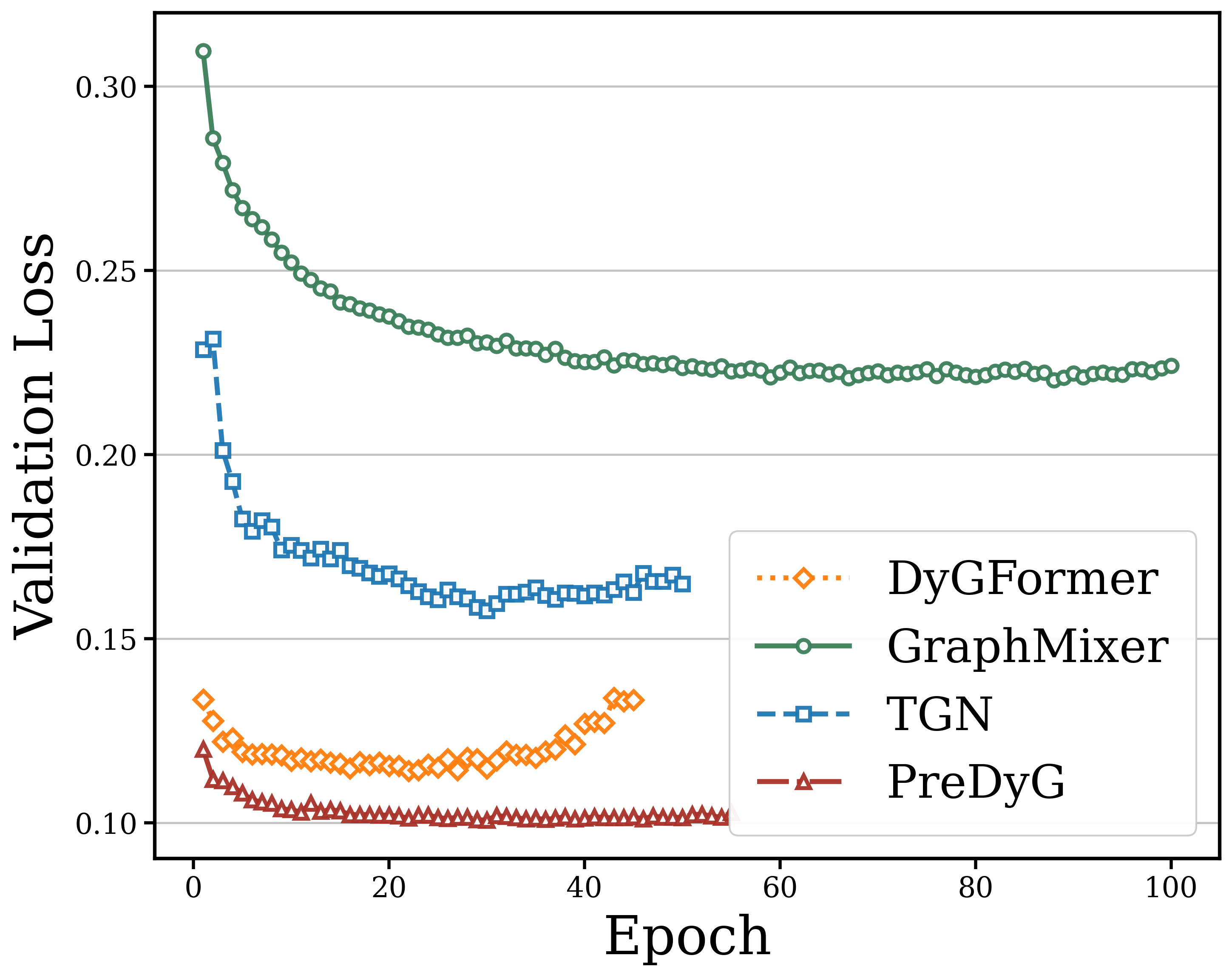}
	\caption{The AP , AUC and loss metrics and generalization gap of multiple training sets with the early stopping mechanism were compared. Training curves comparison across different datasets and evaluation metrics. The first row shows Wikipedia, the second row shows UCI, the third row shows Enron, and the fourth row shows Reddit datasets. From left to right: the average precision (AP) curve, the ROC-AUC curve, and the validation loss curve during training. Each plot displays the PreDyG and other baselines over training time.}
	\label{fig_validation_plots_all}
\end{figure*}

We present additional experimental results that are not fully included, as shown in Figure \ref{fig_validation_plots_all}. The figure provides a detailed analysis of the generalization performance of PreDyG across multiple datasets and visualizes the training process. By tracking validation performance throughout training, we examine the convergence behavior, generalization capability, and differences from baseline methods. The training curves show that PreDyG achieves faster convergence and a more stable training process on all datasets. The validation loss decreases more rapidly and more smoothly, and PreDyG reaches higher peak performance on both AP and AUC. These results further demonstrate that PreDyG not only delivers strong performance but also offers stable training dynamics and robust generalization.
\subsection{Impact of different Negative Sampling Strategies}
\label{app:InfluenceofdifferentNSS}
Tables \ref{tab:all1} to \ref{tab:all4} present a comprehensive comparison of negative sampling strategies for dynamic link prediction. These four tables cover both transductive and inductive evaluation settings. Under each setting, we test three negative sampling strategies, namely random rnd, historical hist, and inductive ind, and evaluate performance using two complementary metrics, AP and AUC-ROC.

Overall, PreDyG achieves the best performance in most experimental configurations, demonstrating a clear advantage. In the transductive setting shown in Tables \ref{tab:all1} and \ref{tab:all2}, PreDyG is particularly strong under random negative sampling, achieving an average rank of 1.00 and outperforming all baselines including TGN, CAWN, and DyGFormer. On \textit{Wikipedia}, PreDyG reaches 99.30±0.01 AP, which is a clear improvement over the second-best method DyGFormer at 98.82±0.02. Under the more challenging historical negative sampling strategy, performance drops for all methods, yet PreDyG remains competitive with an average rank of 2.38, tying with DyGFormer for the best overall rank. Inductive negative sampling is the most challenging setting, but PreDyG still leads on most datasets, with particularly large gains on \textit{Enron}, \textit{LastFM}, and \textit{Contact}.

The inductive setting shown in Tables \ref{tab:all3} and \ref{tab:all4} further confirms the generalization ability and robustness of PreDyG. Under this stricter evaluation protocol, the method continues to deliver strong results and achieves the best performance on almost all datasets under random negative sampling. For example, on \textit{UCI}, the AP of PreDyG decreases slightly from 97.01±0.03 in the transductive setting to 95.75±0.09 in the inductive setting, while still remaining clearly better than competing methods.

Notably, datasets differ substantially in their sensitivity to negative sampling strategies, reflecting the complexity and diversity of real-world dynamic graphs, yet PreDyG remains consistently strong. \textit{Social Evolution} exhibits a relatively stable performance pattern with small gaps among methods, and PreDyG maintains a lead. In contrast, \textit{LastFM} shows larger performance disparities, where the advantage of PreDyG is more pronounced. This consistency across datasets indicates that PreDyG is adaptive and robust, and can effectively handle dynamic graph structures with diverse characteristics.
\input{addition_results}

%% file: tab_tg_data.tex
\begin{table*}[!h]
	\centering
		\caption{The statistical information of each component of the dataset used in the experiment.}
	\small
	\begin{tabular}{c|cc|cc|c|c|c}
		\toprule
		Dataset   & $\text{Dim}_{n}$ & $\text{Dim}_{e}$ & \# Nodes & \# Edges & Information & Duration  &  Time Granularity \\ \midrule
		Wikipedia   & -    & 172  & 9,227    & 157,474  &  Social  & 1 month   &  Unix timestamps  \\ \midrule
		Reddit       & -     & 172   & 10,984   & 672,447  &  Social  & 1 month   &  Unix timestamps   \\ \midrule
				LastFM     & -     & -  & 1,980    & 1,293,103 &  Interaction  & 1 month   &  Unix timestamps   \\ \midrule
		Enron      & -     & -  & 184      & 125,235  &  Communication  & 3 years   &  Unix timestamps   \\ \midrule
		Social Evo. & -     & -  & 59       & 90,879   &  Proximity & 77 days   &  Unix timestamps   \\ \midrule
		Contact    & -     & -  & 274      & 28,244   &  Proximity & 20 days   &  Unix timestamps   \\ \midrule
		Flights    & -     & 4  & 13,169   & 1,470,508 &  Transportation & 3 months &  Unix timestamps   \\ \midrule
		UCI        & -     & -  & 1,899    & 59,835   &  Social & 196 days  &  Unix timestamps   \\ \bottomrule
	\end{tabular}
	\label{tab_tg_data}
\end{table*}

%% file: addition_results.tex
\begin{table*}[!h]
	\centering
	\caption{Average Precision (AP) comparison across different negative sampling strategies (rnd: random, hist: historical, ind: inductive) in transductive link prediction setting.}
	\renewcommand{\arraystretch}{1.1}
	\resizebox{1.0\textwidth}{!}{
		\setlength{\tabcolsep}{0.9mm}{
			\begin{tabular}{c|c|cccccccccc}
				\toprule
				NSS                    & Datasets    & JODIE        & DyRep        & TGAT         & TGN          & CAWN         & EdgeBank     & TCL          & GraphMixer & DyGFormer & PhyDyG    \\ \midrule
				\multirow{9}{*}{rnd}  
				& Wikipedia  & 96.50 $\pm$ 0.18 & 94.86 $\pm$ 0.04 & 96.94 $\pm$ 0.06 & 98.45 $\pm$ 0.05 & 98.76 $\pm$ 0.03 & 90.37 $\pm$ 0.00 & 96.47 $\pm$ 0.14 & 97.25 $\pm$ 0.03 &{98.82 $\pm$ 0.02} &\textbf{99.30$\pm$ 0.01}   \\ 
				& Reddit      & 98.31 $\pm$ 0.12 & 98.22 $\pm$ 0.07 & 98.52 $\pm$ 0.02 & 98.63 $\pm$ 0.05 & {99.11 $\pm$ 0.01} & 94.86 $\pm$ 0.00 & 97.53 $\pm$ 0.02 & 97.31 $\pm$ 0.01 & 99.11 $\pm$ 0.01 &\textbf{99.34 $\pm$ 0.01} \\ 
				& Enron       & 84.77 $\pm$ 0.28 & 82.38 $\pm$ 3.32 & 71.12 $\pm$ 0.96 & 86.53 $\pm$ 1.14 & 89.56 $\pm$ 0.13 & 83.53 $\pm$ 0.00 & 79.70 $\pm$ 0.72 & 82.25 $\pm$ 0.16 & {92.47 $\pm$ 0.15} &\textbf{93.18$\pm$ 0.19} \\ 
				& UCI         & 89.43 $\pm$ 1.06 & 65.14 $\pm$ 2.33 & 79.63 $\pm$ 0.71 & 92.34 $\pm$ 1.02 & 95.18 $\pm$ 0.05 & 76.20 $\pm$ 0.00 & 89.57 $\pm$ 1.61 & 93.25 $\pm$ 0.54 & {95.76 $\pm$ 0.16} &\textbf{97.01$\pm$ 0.03}\\ 
				& LastFM      & 70.85 $\pm$ 2.11 & 71.92 $\pm$ 2.20 & 73.42 $\pm$ 0.18 & 77.07 $\pm$ 3.99 &{86.99 $\pm$ 0.07} & 79.29 $\pm$ 0.00 & 67.27 $\pm$ 2.13 & 75.61 $\pm$ 0.22 & \underline{93.00 $\pm$ 0.10}&\textbf{93.97$\pm$ 0.06}\\ 
				& Social Evo. & 89.89 $\pm$ 0.52 & 88.87 $\pm$ 0.31 & 93.16 $\pm$ 0.20 &{93.57 $\pm$ 0.21} & 84.96 $\pm$ 0.09 & 74.95 $\pm$ 0.00 & 93.13 $\pm$ 0.17 & 93.37 $\pm$ 0.07 & \underline{94.47 $\pm$ 0.01}&\textbf{94.68 $\pm$ 0.01}\\ 
				& Flights     & 95.60 $\pm$ 1.72 & 95.29 $\pm$ 0.69 & 94.03 $\pm$ 0.15 & 97.95 $\pm$ 0.16 & \underline{98.51 $\pm$ 0.01} & 89.35 $\pm$ 0.00 & 91.23 $\pm$ 0.02 & 90.99 $\pm$ 0.08 & {98.25 $\pm$ 0.01}&\textbf{98.79$\pm$ 0.01} \\ 
				& Contact     & 95.31 $\pm$ 1.36 & 95.98 $\pm$ 0.15 & 96.28 $\pm$ 0.06 & {96.89 $\pm$ 0.54} & 90.26 $\pm$ 0.30 & 92.58 $\pm$ 0.00 & 92.44 $\pm$ 0.10 & 91.92 $\pm$ 0.03 & \underline{98.29 $\pm$ 0.01}&\textbf{98.32$\pm$ 0.01}\\ 
		        \cmidrule(lr{0.5em}){2-12}
				& Avg. Rank   &6.50&7.50&6.50&4.00&4.38&8.25&7.88&6.75&\underline{2.25}&\textbf{1.00}   \\ \midrule
				\multirow{9}{*}{hist} 
				& Wikipedia   & 83.01 $\pm$ 0.69 & 79.93 $\pm$ 0.60 & 87.38 $\pm$ 0.19 & 86.86 $\pm$ 0.34 & 71.21 $\pm$ 1.66 & 73.35 $\pm$ 0.00 & \underline{89.05 $\pm$ 0.37} & \textbf{90.90 $\pm$ 0.13} & 82.23 $\pm$ 2.56&85.36$\pm$ 1.17 \\
				& Reddit      & 80.03 $\pm$ 0.31 & 79.83 $\pm$ 0.36 & 79.55 $\pm$ 0.14 & 81.22 $\pm$ 0.59 & 80.82 $\pm$ 0.43 & 73.59 $\pm$ 0.00 & 77.14 $\pm$ 0.18 & 78.44 $\pm$ 0.25 & \underline{81.57 $\pm$ 0.70}&\textbf{82.28$\pm$ 0.47} \\
				& LastFM      & 74.35 $\pm$ 3.82 & 74.92 $\pm$ 2.48 & 71.59 $\pm$ 0.30 & 76.87 $\pm$ 4.60 & 69.86 $\pm$ 0.47 & 73.03 $\pm$ 0.00 & 59.30 $\pm$ 2.34 & 72.47 $\pm$ 0.49 & \underline{81.57 $\pm$ 0.48}&\textbf{83.02$\pm$ 0.49} \\
				& Enron       & 69.85 $\pm$ 2.70 & 71.19 $\pm$ 2.78 & 64.07 $\pm$ 1.04 & 73.91 $\pm$ 1.81 & 64.73 $\pm$ 0.40 & \underline{76.53 $\pm$ 0.00} & 70.66 $\pm$ 0.33 & \textbf{77.98 $\pm$ 0.86} & 75.63 $\pm$ 0.76&75.95$\pm$ 0.39 \\
				& Social Evo. & 87.44 $\pm$ 6.81 & 93.29 $\pm$ 0.42 & 95.01 $\pm$ 0.50 & 94.45 $\pm$ 0.61 & 85.53 $\pm$ 0.41 & 80.57 $\pm$ 0.00 & 94.74 $\pm$ 0.26 & 94.93 $\pm$ 0.25 & \textbf{96.34 $\pm$ 0.17}&\underline{96.87$\pm$ 0.17} \\
				& UCI         & 75.24 $\pm$ 5.80 & 55.10 $\pm$ 3.11 & 68.27 $\pm$ 1.41 & 80.43 $\pm$ 2.09 & 65.30 $\pm$ 0.42 & 65.50 $\pm$ 0.00 & 80.25 $\pm$ 2.75 & \underline{84.11 $\pm$ 1.37} & 82.17 $\pm$ 0.84&\textbf{85.57$\pm$ 0.49} \\
				& Flights     & 66.48 $\pm$ 2.52 & 67.61 $\pm$ 1.01 & \textbf{72.38 $\pm$ 0.12} & 66.70 $\pm$ 1.58 & 64.72 $\pm$ 1.03 & 70.53 $\pm$ 0.00 & 70.68 $\pm$ 0.18 & \underline{71.47 $\pm$ 0.27} & 66.59 $\pm$ 0.53& 66.92$\pm$ 0.54\\
				& Contact     & 95.31 $\pm$ 2.18 & 96.39 $\pm$ 0.19 & 96.05 $\pm$ 0.51 & 93.05 $\pm$ 2.37 & 84.16 $\pm$ 0.52 & 88.81 $\pm$ 0.00 & 93.86 $\pm$ 0.18 & 93.36 $\pm$ 0.41 & \underline{97.57 $\pm$ 0.08}&\textbf{97.92$\pm$ 0.01} \\
				\cmidrule(lr{0.5em}){2-12}
				& Avg. Rank   & 6.50&6.12&5.38&5.00&8.75&7.25&5.88&4.00&\underline{3.75}&\textbf{2.38}     \\ \midrule
				\multirow{9}{*}{ind}  
				& Wikipedia   & 75.65 $\pm$ 0.78 & 70.21 $\pm$ 1.61 & \underline{87.00 $\pm$ 0.16} & 85.62 $\pm$ 0.46 & 74.06 $\pm$ 2.59 & 80.63 $\pm$ 0.00 & 86.76 $\pm$ 0.70 & \textbf{88.59 $\pm$ 0.18} & 78.29 $\pm$ 5.36&80.67$\pm$ 3.15 \\
				& Reddit      & 86.98 $\pm$ 0.15 & 86.30 $\pm$ 0.23 & 89.59 $\pm$ 0.21 & 88.10 $\pm$ 0.22 & \textbf{91.67 $\pm$ 0.26} & 85.48 $\pm$ 0.00 & 87.45 $\pm$ 0.29 & 85.26 $\pm$ 0.12 & 89.12 $\pm$ 0.42&\underline{90.18$\pm$ 0.18} \\
				& LastFM      & 62.67 $\pm$ 4.50 & 64.41 $\pm$ 2.73 & 71.13 $\pm$ 0.14 & 65.95 $\pm$ 6.01 & 67.48 $\pm$ 0.78 & \underline{75.49 $\pm$ 0.00} & 58.21 $\pm$ 0.88 & 68.12 $\pm$ 0.35 & 73.97 $\pm$ 0.46& \textbf{77.73$\pm$ 0.64}\\
				& Enron       & 68.96 $\pm$ 0.98 & 67.79 $\pm$ 1.53 & 63.94 $\pm$ 1.34 & 70.89 $\pm$ 2.70 & 75.15 $\pm$ 0.61 & 73.89 $\pm$ 0.00 & 71.29 $\pm$ 0.30 & 75.01 $\pm$ 0.76 & \underline{77.41 $\pm$ 0.92}&\textbf{81.74$\pm$ 0.49} \\
				& Social Evo. & 89.82 $\pm$ 4.12 & 93.28 $\pm$ 0.46 & 94.84 $\pm$ 0.41 & 95.13 $\pm$ 0.60 & 88.32 $\pm$ 0.29 & 83.69 $\pm$ 0.00 & 94.90 $\pm$ 0.36 & 94.72 $\pm$ 0.32 & \underline{95.59 $\pm$ 0.10}&\textbf{95.72$\pm$ 0.12} \\
				& UCI         & 65.99 $\pm$ 1.40 & 54.79 $\pm$ 1.73 & 68.67 $\pm$ 0.86 & 70.94 $\pm$ 0.74 & 64.61 $\pm$ 0.46 & 57.43 $\pm$ 0.00 & 76.01 $\pm$ 1.07 & \underline{80.10 $\pm$ 0.51} & 72.25 $\pm$ 1.68&\textbf{83.52$\pm$ 0.56} \\
				& Flights     & 69.07 $\pm$ 4.03 & 70.57 $\pm$ 1.81 & \underline{75.48 $\pm$ 0.23} & 71.09 $\pm$ 2.75 & 69.18 $\pm$ 1.54 & \textbf{81.08 $\pm$ 0.00} & 74.62 $\pm$ 0.16 & 74.87 $\pm$ 0.24 & 70.92 $\pm$ 1.74& 70.75$\pm$ 0.98\\
				& Contact     & 93.43 $\pm$ 1.74 & 94.18 $\pm$ 0.12 & 94.35 $\pm$ 0.47 & 90.18 $\pm$ 3.26 & 89.31 $\pm$ 0.25 & 85.20 $\pm$ 0.00 & 91.35 $\pm$ 0.19 & 90.87 $\pm$ 0.32 & \underline{94.75 $\pm$ 0.31}&\textbf{95.85$\pm$ 0.23} \\
				\cmidrule(lr{0.5em}){2-12}
				& Avg. Rank  &7.75&8.00&4.38&5.50&6.75&6.50&5.25&4.75&\underline{3.75}&\textbf{2.38} \\ \bottomrule
			\end{tabular}
		}
	}
	\label{tab:all1}
\end{table*}
\begin{table*}[!h]
	\centering
	\caption{AUC-ROC comparison across different negative sampling strategies in transductive link prediction setting.}
	\renewcommand{\arraystretch}{1.1}
	\resizebox{1.0\textwidth}{!}{
		\setlength{\tabcolsep}{0.9mm}{
			\begin{tabular}{c|c|cccccccccc}
				\toprule
				NSS                    & Datasets    & JODIE        & DyRep        & TGAT         & TGN          & CAWN         & EdgeBank     & TCL          & GraphMixer & DyGFormer & PhyDyG  \\ \midrule
				\multirow{9}{*}{rnd}  
				& Wikipedia   & 96.33 $\pm$ 0.05 & 94.37 $\pm$ 0.06 & 96.67 $\pm$ 0.07 & 98.37 $\pm$ 0.07 & 98.54 $\pm$ 0.04 & 90.78 $\pm$ 0.00 & 95.84 $\pm$ 0.15 & 96.92 $\pm$ 0.03 & \underline{98.91 $\pm$ 0.02} &\textbf{99.30$\pm$ 0.01}\\
				& Reddit      & 98.31 $\pm$ 0.06 & 98.17 $\pm$ 0.02 & 98.47 $\pm$ 0.02 & 98.60 $\pm$ 0.07 & 99.01 $\pm$ 0.01 & 95.37 $\pm$ 0.00 & 97.42 $\pm$ 0.02 & 97.17 $\pm$ 0.02 & \underline{99.15 $\pm$ 0.01} &\textbf{99.31$\pm$ 0.01}\\
				& LastFM      & 70.49 $\pm$ 1.70 & 71.16 $\pm$ 1.86 & 71.59 $\pm$ 0.17 & 78.47 $\pm$ 2.94 & 85.92 $\pm$ 0.09 & 83.77 $\pm$ 0.00 & 64.06 $\pm$ 1.17 & 73.53 $\pm$ 0.09 & \underline{93.05 $\pm$ 0.10} &\textbf{94.10$\pm$ 0.07}\\
				& Enron       & 87.96 $\pm$ 0.51 & 84.89 $\pm$ 3.04 & 68.89 $\pm$ 1.08 & 88.32 $\pm$ 0.95 & 90.45 $\pm$ 0.17 & 87.05 $\pm$ 0.00 & 75.74 $\pm$ 0.73 & 84.38 $\pm$ 0.24 & \underline{93.33 $\pm$ 0.14}&\textbf{93.88$\pm$ 0.02}\\
				& Social Evo. & 92.05 $\pm$ 0.47 & 90.76 $\pm$ 0.22 & 94.76 $\pm$ 0.16 & 95.39 $\pm$ 0.13 & 87.34 $\pm$ 0.10 & 81.60 $\pm$ 0.00 & 94.84 $\pm$ 0.20 & 95.23 $\pm$ 0.04 & \underline{96.30 $\pm$ 0.01}&\textbf{96.40$\pm$ 0.01}\\
				& UCI         & 90.44 $\pm$ 0.45 & 68.77 $\pm$ 2.38 & 78.53 $\pm$ 0.71 & 92.03 $\pm$ 1.11 & 93.87 $\pm$ 0.11 & 77.30 $\pm$ 0.00 & 87.82 $\pm$ 1.39 & 91.81 $\pm$ 0.69 & \underline{94.49 $\pm$ 0.29}&\textbf{96.22$\pm$ 0.03}\\
				& Flights     & 96.21 $\pm$ 1.42 & 95.95 $\pm$ 0.60 & 94.13 $\pm$ 0.16 & 98.22 $\pm$ 0.10 & 98.45 $\pm$ 0.01 & 90.23 $\pm$ 0.00 & 91.21 $\pm$ 0.02 & 91.13 $\pm$ 0.01 & \textbf{98.53 $\pm$ 0.01}&\underline{98.86$\pm$ 0.01} \\
				& Contact     & 96.66 $\pm$ 0.87 & 96.48 $\pm$ 0.14 & 96.95 $\pm$ 0.05 & 97.54 $\pm$ 0.37 & 89.99 $\pm$ 0.36 & 94.34 $\pm$ 0.00 & 94.15 $\pm$ 0.05 & 93.94 $\pm$ 0.02 & \underline{98.53 $\pm$ 0.01}&\textbf{98.60$\pm$ 0.01}\\ 
				\cmidrule(lr{0.5em}){2-12} 
				& Avg. Rank   & 6.25&7.62&6.62&3.88&4.62&8.25&7.88&6.88&\underline{2.00}&\textbf{1.00}\\ \midrule
				\multirow{9}{*}{hist} 
				& Wikipedia   & 80.77 $\pm$ 0.75 & 77.74 $\pm$ 0.32 & 82.87 $\pm$ 0.26 & 82.74 $\pm$ 0.31 & 67.84 $\pm$ 0.63 & 77.27 $\pm$ 0.00 & \underline{85.76 $\pm$ 0.43} & \textbf{87.68 $\pm$ 0.18} & 78.80 $\pm$ 1.94&83.37$\pm$ 0.52 \\
				& Reddit      & 80.52 $\pm$ 0.35 & 80.15 $\pm$ 0.21 & 79.33 $\pm$ 0.18 & \underline{81.11 $\pm$ 0.22} & 80.27 $\pm$ 0.28 & 78.58 $\pm$ 0.00 & 76.49 $\pm$ 0.15 & 77.80 $\pm$ 0.10 & 80.54 $\pm$ 0.27&\textbf{81.79$\pm$ 0.22} \\
				& LastFM      & 75.22 $\pm$ 2.33 & 74.65 $\pm$ 2.02 & 64.27 $\pm$ 0.23 & 77.97 $\pm$ 3.00 & 67.88 $\pm$ 0.22 & 78.09 $\pm$ 0.00 & 47.24 $\pm$ 3.12 & 64.21 $\pm$ 0.74 & \underline{78.78 $\pm$ 0.33}&\textbf{79.68$\pm$ 0.29} \\
				& Enron       & 75.39 $\pm$ 2.39 & 74.69 $\pm$ 3.55 & 61.85 $\pm$ 1.45 & \underline{77.09 $\pm$ 2.22} & 65.10 $\pm$ 0.34 & \textbf{79.59 $\pm$ 0.00} & 67.95 $\pm$ 0.92 & 75.27 $\pm$ 1.11 & 76.55 $\pm$ 0.51&74.65$\pm$ 0.30\\
				& Social Evo. & 90.06 $\pm$ 3.14 & 93.12 $\pm$ 0.36 & 93.08 $\pm$ 0.58 & 94.71 $\pm$ 0.49 & 87.43 $\pm$ 0.17 & 85.81 $\pm$ 0.00 & 93.44 $\pm$ 0.70 & 94.39$\pm$ 0.34& \underline{96.44$\pm$ 0.10}&\textbf{97.00$\pm$ 0.15}\\
				& UCI         & \underline{78.64 $\pm$ 3.52} & 57.91 $\pm$ 3.10 & 58.89 $\pm$ 1.60 & 77.25 $\pm$ 2.71 & 57.86 $\pm$ 0.13 & 69.56 $\pm$ 0.00 & 72.25 $\pm$ 3.42 & 77.54 $\pm$ 2.04 & 76.97 $\pm$ 0.22& \textbf{79.52$\pm$ 0.29} \\
				& Flights     & 68.97 $\pm$ 1.88 & 69.43 $\pm$ 0.87 & \underline{72.20 $\pm$ 0.15} & 68.39 $\pm$ 0.96 & 66.11 $\pm$ 0.75 & \textbf{74.64 $\pm$ 0.00} & 70.57 $\pm$ 0.19 & 70.37 $\pm$ 0.21 & 68.09 $\pm$ 0.44&69.73$\pm$ 0.32 \\
				& Contact     & 96.35 $\pm$ 0.89 & 96.00 $\pm$ 0.23 & 95.39 $\pm$ 0.41 & 93.76 $\pm$ 1.27 & 83.06 $\pm$ 0.33 & 92.17 $\pm$ 0.00 & 93.34 $\pm$ 0.20 & 93.14 $\pm$ 0.32 & \underline{97.17 $\pm$ 0.08}&\textbf{97.41$\pm$ 0.07} \\
				\cmidrule(lr{0.5em}){2-12}
				& Avg. Rank   & 4.88&6.38&6.38&4.25&8.75&6.00&6.38&5.38&\underline{4.12}&\textbf{2.50} \\ \midrule
				\multirow{9}{*}{ind}  
				& Wikipedia   & 70.96 $\pm$ 0.75 & 67.36 $\pm$ 0.94 & 81.93 $\pm$ 0.25 & 80.97 $\pm$ 0.33 & 70.97 $\pm$ 0.97 & 81.73 $\pm$ 0.00 & \underline{82.19 $\pm$ 0.47} & \textbf{84.28 $\pm$ 0.30} & 75.09 $\pm$ 3.68&81.64$\pm$ 1.91 \\
				& Reddit      & 83.51 $\pm$ 0.12 & 82.90 $\pm$ 0.28 & 87.13 $\pm$ 0.18 & 84.56 $\pm$ 0.26 & \textbf{88.04 $\pm$ 0.32} & 85.93 $\pm$ 0.00 & 84.67 $\pm$ 0.26 & 82.21 $\pm$ 0.12 & 86.23 $\pm$ 0.53&\underline{87.18$\pm$ 0.57} \\
				& LastFM      & 61.32 $\pm$ 3.51 & 62.15 $\pm$ 2.14 & 63.99 $\pm$ 0.18 & 65.46 $\pm$ 4.24 & 67.92 $\pm$ 0.40 & \textbf{77.37 $\pm$ 0.00} & 46.93 $\pm$ 2.56 & 60.22 $\pm$ 0.33 & 69.25 $\pm$ 0.40&\underline{70.36$\pm$ 0.62} \\
				& Enron       & 70.92 $\pm$ 1.06 & 68.73 $\pm$ 1.34 & 60.45 $\pm$ 2.11 & 71.34 $\pm$ 2.50 & \underline{75.17 $\pm$ 0.51} & 75.00 $\pm$ 0.00 & 67.64 $\pm$ 0.87 & 71.53 $\pm$ 0.82 & 74.07 $\pm$ 0.66&\textbf{81.06$\pm$ 0.23} \\
				& Social Evo. & 90.01 $\pm$ 3.20 & 93.07 $\pm$ 0.38 & 92.94 $\pm$ 0.59 & 95.24 $\pm$ 0.55 & 89.93 $\pm$ 0.12 & 87.88 $\pm$ 0.00 & 93.44 $\pm$ 0.71 & 94.22 $\pm$ 0.31 & \underline{96.22 $\pm$ 0.09}&\textbf{96.37$\pm$ 0.03} \\
				& UCI         & 64.14 $\pm$ 1.28 & 54.25 $\pm$ 2.01 & 60.80 $\pm$ 1.03 & 64.11 $\pm$ 1.07 & 58.06 $\pm$ 0.25 & 58.03 $\pm$ 0.00 & 70.05 $\pm$ 1.85 & \underline{74.59 $\pm$ 0.76} & 65.96 $\pm$ 1.21&\textbf{75.23$\pm$ 0.52} \\
				& Flights     & 69.99 $\pm$ 3.13 & 71.13 $\pm$ 1.58 & \underline{73.47 $\pm$ 0.20} & 71.63 $\pm$ 1.71 & 69.70 $\pm$ 0.73 & \textbf{81.10 $\pm$ 0.00} & 72.54 $\pm$ 0.16 & 72.21 $\pm$ 0.20 & 69.53 $\pm$ 1.17& 71.23$\pm$ 0.82\\
				& Contact     & 94.47 $\pm$ 1.10 & 94.23 $\pm$ 0.15 & 94.10 $\pm$ 0.39 & 91.64 $\pm$ 1.68 & 87.68 $\pm$ 0.23 & 85.87 $\pm$ 0.00 & 91.23 $\pm$ 0.19 & 90.96 $\pm$ 0.25 & \underline{95.01 $\pm$ 0.15}&\textbf{95.47$\pm$ 0.09}\\ 
				\cmidrule(lr{0.5em}){2-12}
				& Avg. Rank   &7.00&7.62&5.38&5.50&6.25&5.38&5.62&5.38&\underline{4.50}&\textbf{2.38} \\ \bottomrule
			\end{tabular}
		}
	}
	\label{tab:all2}
\end{table*}			
\begin{table*}[!h]
	\centering
	\caption{Average Precision (AP) comparison across different negative sampling strategies in inductive link prediction setting.}
	\renewcommand{\arraystretch}{1.1}
	\resizebox{1.0\textwidth}{!}{
		\setlength{\tabcolsep}{0.9mm}{
			\begin{tabular}{c|c|ccccccccc}
				\toprule
				NSS                    & Datasets    & JODIE        & DyRep        & TGAT         & TGN          & CAWN         & TCL          & GraphMixer & DyGFormer & PhyDyG   \\ \midrule
				\multirow{9}{*}{rnd}  
				& Wikipedia   & 94.82 $\pm$ 0.23 & 92.43 $\pm$ 0.38 & 96.22 $\pm$ 0.06 & 97.83 $\pm$ 0.07 & 98.24 $\pm$ 0.03 & 96.22 $\pm$ 0.15 & 96.65 $\pm$ 0.02 & \underline{98.59 $\pm$ 0.03}&\textbf{98.96$\pm$ 0.01} \\
				& Reddit      & 96.50 $\pm$ 0.16 & 96.09 $\pm$ 0.11 & 97.09 $\pm$ 0.06 & 97.50 $\pm$ 0.09 & 98.62 $\pm$ 0.01 & 94.09 $\pm$ 0.11 & 95.26 $\pm$ 0.02 & \underline{98.84 $\pm$ 0.02}&\textbf{99.00$\pm$ 0.01} \\
				& LastFM      & 81.61 $\pm$ 3.81 & 83.02 $\pm$ 1.44 & 78.63 $\pm$ 0.34 & 81.45 $\pm$ 4.29 & 89.42 $\pm$ 0.06 & 73.53 $\pm$ 1.67 & 82.11 $\pm$ 0.40 & \underline{94.23 $\pm$ 0.06}&\textbf{95.18$\pm$ 0.02} \\
				& Enron       & 80.72 $\pm$ 1.42 & 74.55 $\pm$ 3.93 & 67.05 $\pm$ 1.54 & 77.94 $\pm$ 1.05 & 86.35 $\pm$ 0.51 & 76.14 $\pm$ 0.82 & 75.88 $\pm$ 0.46 & \underline{89.76 $\pm$ 0.34}&\textbf{89.90$\pm$ 0.19} \\
				& Social Evo. & 91.96 $\pm$ 0.49 & 90.04 $\pm$ 0.50 & 91.41 $\pm$ 0.13 & 90.77 $\pm$ 0.89 & 79.94 $\pm$ 0.20 & 91.55 $\pm$ 0.08 & 91.86 $\pm$ 0.06 & \underline{93.14 $\pm$ 0.06}&\textbf{93.20$\pm$ 0.08}\\
				& UCI         & 79.86 $\pm$ 1.46 & 57.48 $\pm$ 1.85 & 79.54 $\pm$ 0.51 & 88.12 $\pm$ 2.05 & 92.73 $\pm$ 0.07 & 87.36 $\pm$ 2.04 & 91.19 $\pm$ 0.41 & \underline{94.54 $\pm$ 0.15}&\textbf{95.75$\pm$ 0.09}\\
				& Flights     & 94.74 $\pm$ 0.41 & 92.88 $\pm$ 0.72 & 88.73 $\pm$ 0.33 & 95.03 $\pm$ 0.64 & 97.06 $\pm$ 0.02 & 83.41 $\pm$ 0.10 & 83.03 $\pm$ 0.05 & \textbf{97.79 $\pm$ 0.02}&\underline{97.54$\pm$ 0.01} \\
				& Contact     & 94.34 $\pm$ 1.46 & 92.18 $\pm$ 0.42 & 95.87 $\pm$ 0.07 & 93.82 $\pm$ 1.02 & 89.55 $\pm$ 0.27 & 91.11 $\pm$ 0.09 & 90.59 $\pm$ 0.08 & \underline{98.03 $\pm$ 0.02}&\textbf{98.07$\pm$ 0.01} \\
				\cmidrule(lr{0.5em}){2-11}
				& Avg. Rank   & 5.38&7.12&6.50&5.12&4.50&7.12&6.25&\underline{1.88}&\textbf{1.12}\\ \midrule
				\multirow{9}{*}{hist} 
				& Wikipedia   & 68.69 $\pm$ 0.37 & 62.18 $\pm$ 1.24 & \underline{84.17 $\pm$ 0.21} & 81.76 $\pm$ 0.36 & 67.27 $\pm$ 1.66 & 82.20 $\pm$ 2.16 & \textbf{87.60 $\pm$ 0.28} & 71.42 $\pm$ 4.44&76.41$\pm$ 3.16 \\
				& Reddit      & 62.34 $\pm$ 0.50 & 61.60 $\pm$ 0.73 & 63.47 $\pm$ 0.36 & 64.85 $\pm$ 0.84 & 63.67 $\pm$ 0.42 & 60.83 $\pm$ 0.29 & 64.50 $\pm$ 0.23 & \underline{65.37 $\pm$ 0.59}&\textbf{67.25$\pm$ 0.69} \\
				& LastFM      & 70.39 $\pm$ 4.29 & 71.45 $\pm$ 1.76 & 76.27 $\pm$ 0.28 & 66.65 $\pm$ 6.09 & 71.33 $\pm$ 0.47 & 65.78 $\pm$ 0.63 & \underline{76.42 $\pm$ 0.25} & 76.35 $\pm$ 0.52& \textbf{77.73$\pm$ 0.47}\\
				& Enron       & 65.86 $\pm$ 3.73 & 62.08 $\pm$ 2.24 & 61.40 $\pm$ 1.34 & 62.91 $\pm$ 1.16 & 60.70 $\pm$ 0.32 & 67.11 $\pm$ 0.63 & \underline{72.37 $\pm$ 1.40} & 67.07 $\pm$ 0.65&\textbf{81.74$\pm$ 0.39} \\
				& Social Evo. & 88.51 $\pm$ 0.91 & 88.72 $\pm$ 1.06 & 93.97 $\pm$ 0.50 & 90.66 $\pm$ 1.65 & 79.83 $\pm$ 0.37 & 94.10 $\pm$ 0.33 & 94.01 $\pm$ 0.43 & \underline{94.97 $\pm$ 0.24}&\textbf{95.73$\pm$ 0.25} \\
				& UCI         & 63.11 $\pm$ 2.29 & 52.47 $\pm$ 2.05 & 70.52 $\pm$ 0.97 & 70.80 $\pm$ 0.80 & 64.54 $\pm$ 0.45 & 76.71 $\pm$ 0.98 & \underline{81.66 $\pm$ 0.47} & 72.13 $\pm$ 1.90&\textbf{83.52$\pm$ 0.67} \\
				& Flights     & 61.01 $\pm$ 1.66 & 62.83 $\pm$ 1.31 & \underline{64.72 $\pm$ 0.38} & 59.31 $\pm$ 1.45 & 56.82 $\pm$ 0.55 & 64.50 $\pm$ 0.23 & \textbf{65.28 $\pm$ 0.28} & 57.11 $\pm$ 0.25& 57.77$\pm$ 0.33\\
				& Contact     & 90.42 $\pm$ 2.37 & 89.22 $\pm$ 0.69 & \underline{94.15 $\pm$ 0.43} & 88.13 $\pm$ 1.48 & 74.20 $\pm$ 0.79 & 90.44 $\pm$ 0.16 & 89.91 $\pm$ 0.40 & 93.56 $\pm$ 0.54&\textbf{95.20$\pm$ 0.47} \\ 
				\cmidrule(lr{0.5em}){2-11}
				& Avg. Rank   & 6.50&7.00&4.38&5.75&7.75&4.62&\underline{2.75}&4.00&\textbf{2.25}\\ \midrule
				\multirow{9}{*}{ind}  
				& Wikipedia   & 68.70 $\pm$ 0.36 & 62.19 $\pm$ 1.26 & \underline{84.17 $\pm$ 0.19} & 81.77 $\pm$ 0.32 & 67.24 $\pm$ 1.66 & 82.20 $\pm$ 2.21 & \textbf{87.60 $\pm$ 0.25} & 71.42 $\pm$ 4.45&76.42$\pm$ 1.15 \\
				& Reddit      & 62.32 $\pm$ 0.51 & 61.58 $\pm$ 0.70 & 63.40 $\pm$ 0.40 & 64.84 $\pm$ 0.83 & 63.65 $\pm$ 0.39 & 60.81 $\pm$ 0.22 & 64.49 $\pm$ 0.26 & \underline{65.35 $\pm$ 0.63}&\textbf{67.30$\pm$ 0.49} \\
				& LastFM      & 70.39 $\pm$ 4.33 & 71.45 $\pm$ 1.71 & 76.28 $\pm$ 0.21 & 69.46 $\pm$ 4.64 & 71.33 $\pm$ 0.44 & 65.78 $\pm$ 0.64 & \underline{76.42 $\pm$ 0.22} & 76.35 $\pm$ 0.50& \textbf{77.73$\pm$ 0.50}\\
				& Enron       & 65.86 $\pm$ 3.68 & 62.08 $\pm$ 2.25 & 61.40 $\pm$ 1.30 & 62.90 $\pm$ 1.19 & 60.72 $\pm$ 0.37 & 67.11 $\pm$ 0.60 & \underline{72.37 $\pm$ 1.40} & 67.07 $\pm$ 0.62&\textbf{81.74$\pm$ 0.34} \\
				& Social Evo. & 88.51 $\pm$ 0.87 & 88.72 $\pm$ 1.11 & 93.97 $\pm$ 0.53 & 90.65 $\pm$ 1.58 & 79.83 $\pm$ 0.37 & 94.10 $\pm$ 0.35 & 94.01 $\pm$ 0.45 & \underline{95.11 $\pm$ 0.22}&\textbf{95.72$\pm$ 0.28} \\
				& UCI         & 63.16 $\pm$ 2.29 & 52.47 $\pm$ 2.08 & 70.49 $\pm$ 0.95 & 70.73 $\pm$ 0.82 & 64.54 $\pm$ 0.49 & 76.65 $\pm$ 1.02 & \underline{81.64 $\pm$ 0.45} & 72.13 $\pm$ 1.90&\textbf{83.52$\pm$ 0.75} \\
				& Flights     & 61.01 $\pm$ 1.64 & 62.83 $\pm$ 1.30 & \underline{64.72 $\pm$ 0.36} & 59.32 $\pm$ 1.44 & 56.82 $\pm$ 0.60 & 64.50 $\pm$ 0.23 & \textbf{65.29 $\pm$ 0.26} & 57.11 $\pm$ 0.22& 57.81$\pm$ 0.18\\
				& Contact     & 90.43 $\pm$ 2.35 & 89.22 $\pm$ 0.69 & \underline{94.14 $\pm$ 0.46} & 88.12 $\pm$ 1.46 & 74.19 $\pm$ 0.80 & 90.43 $\pm$ 0.17 & 89.91 $\pm$ 0.39 & 93.55 $\pm$ 0.48&\textbf{95.21$\pm$ 0.44} \\ 
				\cmidrule(lr{0.5em}){2-11}
				& Avg. Rank   &6.38&7.00&4.38&5.75&7.75&4.75&\underline{2.75}&4.00&\textbf{2.25}  \\ \bottomrule
			\end{tabular}
		}
	}
	\label{tab:all3}
\end{table*}
\begin{table*}[!h]
	\centering
	\caption{AUC-ROC comparison across different negative sampling strategies in inductive link prediction setting.}
	\renewcommand{\arraystretch}{1.1}
	\resizebox{1.0\textwidth}{!}{
		\setlength{\tabcolsep}{0.9mm}{
			\begin{tabular}{c|c|ccccccccc}
				\toprule
				NSS                    & Datasets    & JODIE        & DyRep        & TGAT         & TGN          & CAWN         & TCL          & GraphMixer & DyGFormer & PhyDyG   \\ \midrule
				\multirow{9}{*}{rnd}  
				& Wikipedia   & 94.33 $\pm$ 0.28 & 91.49 $\pm$ 0.47 & 95.90 $\pm$ 0.10 & 97.72 $\pm$ 0.03 & 98.03 $\pm$ 0.03 & 95.57 $\pm$ 0.18 & 96.30 $\pm$ 0.06 & \underline{98.48 $\pm$ 0.03}&\textbf{98.93$\pm$ 0.01} \\
				& Reddit      & 96.52 $\pm$ 0.10 & 96.05 $\pm$ 0.11 & 96.98 $\pm$ 0.07 & 97.39 $\pm$ 0.08 & 98.42 $\pm$ 0.02 & 93.80 $\pm$ 0.03 & 94.97 $\pm$ 0.02 & \underline{98.71 $\pm$ 0.01}&\textbf{98.93$\pm$ 0.01}\\
				& LastFM      & 81.13 $\pm$ 3.37 & 82.24 $\pm$ 1.48 & 76.99 $\pm$ 0.26 & 82.61 $\pm$ 3.15 & 87.82 $\pm$ 0.15 & 70.84 $\pm$ 0.82 & 80.37 $\pm$ 0.21 & \underline{94.08 $\pm$ 0.07}&\textbf{95.12$\pm$ 0.02}\\
				& Enron       & 81.96 $\pm$ 1.37 & 76.34 $\pm$ 4.20 & 64.63 $\pm$ 1.72 & 78.83 $\pm$ 1.13 & 87.02 $\pm$ 0.47 & 72.33 $\pm$ 0.96 & 76.51 $\pm$ 0.71 & \underline{90.69 $\pm$ 0.29}&\textbf{90.84$\pm$ 0.17} \\
				& Social Evo. & 93.70 $\pm$ 0.32 & 91.18 $\pm$ 0.52 & 93.41 $\pm$ 0.17 & 93.43 $\pm$ 0.57 & 84.73 $\pm$ 0.30 & 93.71 $\pm$ 0.17 & 94.09 $\pm$ 0.10 & \underline{95.29 $\pm$ 0.03}&\textbf{95.37$\pm$ 0.05}\\
				& UCI         & 78.80 $\pm$ 0.97 & 58.08 $\pm$ 1.79 & 77.64 $\pm$ 0.41 & 86.68 $\pm$ 2.26 & 90.40 $\pm$ 0.13 & 84.49 $\pm$ 1.84 & 89.30 $\pm$ 0.59 & \underline{92.63 $\pm$ 0.10}&\textbf{94.40$\pm$ 0.13} \\
				& Flights     & 95.21 $\pm$ 0.33 & 93.56 $\pm$ 0.73 & 88.64 $\pm$ 0.32 & 95.92 $\pm$ 0.40 & 96.86 $\pm$ 0.02 & 82.48 $\pm$ 0.01 & 82.27 $\pm$ 0.03 & \textbf{97.80 $\pm$ 0.02}&\underline{97.64$\pm$ 0.03}  \\
				& Contact     & 95.37 $\pm$ 0.93 & 91.89 $\pm$ 0.41 & 96.53 $\pm$ 0.14 & 94.84 $\pm$ 0.71 & 89.07 $\pm$ 0.30 & 93.05 $\pm$ 0.11 & 92.83 $\pm$ 0.02 & \underline{98.30 $\pm$ 0.02}&\textbf{98.37$\pm$ 0.03}\\
				\cmidrule(lr{0.5em}){2-11} 
				& Avg. Rank   & 5.62&7.38&6.62&4.62&4.50&7.12&6.12&\underline{1.88}&\textbf{1.12}\\ \midrule
				\multirow{9}{*}{hist} 
				& Wikipedia   & 61.86 $\pm$ 0.51 & 57.54 $\pm$ 1.12 & 78.38 $\pm$ 0.16 & 75.75 $\pm$ 0.28 & 62.04 $\pm$ 0.63 & \underline{79.79 $\pm$ 0.97} & \textbf{82.87 $\pm$ 0.17} & 68.33 $\pm$ 2.82&75.23$\pm$ 2.03 \\
				& Reddit      & 61.69 $\pm$ 0.40 & 60.45 $\pm$ 0.40 & 64.43 $\pm$ 0.30 & 64.55 $\pm$ 0.47 & \underline{64.94 $\pm$ 0.20} & 61.43 $\pm$ 0.26 & 64.27 $\pm$ 0.16 & 64.81 $\pm$ 0.24&\textbf{67.14$\pm$ 0.41} \\
				& LastFM      & 68.44 $\pm$ 3.25 & 68.79 $\pm$ 1.08 & 69.89 $\pm$ 0.24 & 66.99 $\pm$ 5.65 & 67.69 $\pm$ 0.23 & 55.88 $\pm$ 1.84 & 70.07 $\pm$ 0.22 & \textbf{70.73 $\pm$ 0.33}&\underline{70.36$\pm$ 0.29}\\
				& Enron       & 65.32 $\pm$ 3.56 & 61.50 $\pm$ 2.52 & 57.84 $\pm$ 2.18 & 62.68 $\pm$ 1.06 & 62.25 $\pm$ 0.42 & 64.06 $\pm$ 1.05 & \underline{68.20 $\pm$ 1.62} & 65.78 $\pm$ 0.44&\textbf{81.06$\pm$ 0.48} \\
				& Social Evo. & 88.53 $\pm$ 0.53 & 87.93 $\pm$ 1.04 & 91.87 $\pm$ 0.69 & 92.10 $\pm$ 1.26 & 83.54 $\pm$ 0.21 & 93.28 $\pm$ 0.59 & 93.62 $\pm$ 0.33 & \underline{96.12 $\pm$ 0.08}&\textbf{96.37$\pm$ 0.08} \\
				& UCI         & 60.24 $\pm$ 1.95 & 51.25 $\pm$ 2.35 & 62.32 $\pm$ 1.14 & 62.69 $\pm$ 0.92 & 56.39 $\pm$ 0.08 & 70.46 $\pm$ 1.90 & \textbf{75.98 $\pm$ 0.80} & 65.55 $\pm$ 1.01&\underline{75.23$\pm$ 0.35} \\
				& Flights     & 60.72 $\pm$ 1.31 & 61.99 $\pm$ 1.42 & \underline{63.38 $\pm$ 0.29} & 59.66 $\pm$ 1.03 & 56.58 $\pm$ 0.46 & \textbf{63.48 $\pm$ 0.24} & 63.30 $\pm$ 0.19 & 56.05 $\pm$ 0.22& 61.25$\pm$ 0.39\\
				& Contact     & 90.80 $\pm$ 1.16 & 88.88 $\pm$ 0.64 & 93.76 $\pm$ 0.40 & 88.84 $\pm$ 1.43 & 74.79 $\pm$ 0.34 & 90.37 $\pm$ 0.19 & 90.04 $\pm$ 0.30 & \underline{94.14 $\pm$ 0.27}&\textbf{95.48$\pm$ 0.24}\\  \cmidrule(lr{0.5em}){2-11}
				& Avg. Rank   & 6.12&7.38&4.75&5.88&7.12&4.62&\underline{3.12}&3.75&\textbf{2.25}\\ \midrule
				\multirow{9}{*}{ind}  
				& Wikipedia   & 61.87 $\pm$ 0.49 & 57.54 $\pm$ 1.07 & 78.38 $\pm$ 0.22 & 75.76 $\pm$ 0.31 & 62.02 $\pm$ 0.66 & \underline{79.79 $\pm$ 0.94} & \textbf{82.88 $\pm$ 0.23} & 68.33 $\pm$ 2.84&76.24$\pm$ 1.43 \\
				& Reddit      & 61.69 $\pm$ 0.36 & 60.44 $\pm$ 0.39 & 64.39 $\pm$ 0.23 & 64.55 $\pm$ 0.49 & \underline{64.91 $\pm$ 0.21} & 61.36 $\pm$ 0.27 & 64.27 $\pm$ 0.14 & 64.80 $\pm$ 0.24&\textbf{67.18$\pm$ 0.25} \\
				& LastFM      & 68.44 $\pm$ 3.28 & 68.79 $\pm$ 1.08 & 69.89 $\pm$ 0.29 & 66.99 $\pm$ 5.59 & 67.68 $\pm$ 0.24 & 55.88 $\pm$ 1.88 & 70.07 $\pm$ 0.24 & \textbf{70.73 $\pm$ 0.33}& \underline{70.36$\pm$ 0.36}\\
				& Enron       & 65.32 $\pm$ 3.60 & 61.50 $\pm$ 2.49 & 57.83 $\pm$ 2.17 & 62.68 $\pm$ 1.08 & 62.27 $\pm$ 0.39 & 64.05 $\pm$ 1.03 & \underline{68.19 $\pm$ 1.61} & 65.79 $\pm$ 0.38&\textbf{81.06$\pm$ 0.30} \\
				& Social Evo. & 88.53 $\pm$ 0.54 & 87.93 $\pm$ 1.05 & 91.88 $\pm$ 0.74 & 92.10 $\pm$ 1.25 & 83.54 $\pm$ 0.23 & 93.28 $\pm$ 0.62 & 93.62 $\pm$ 0.35 & \underline{96.03 $\pm$ 0.05}&\textbf{96.37$\pm$ 0.05} \\
				& UCI         & 60.27 $\pm$ 1.94 & 51.26 $\pm$ 2.38 & 62.29 $\pm$ 1.21 & 62.66 $\pm$ 0.88 & 56.39 $\pm$ 0.14 & 70.42 $\pm$ 1.96 & \textbf{75.97 $\pm$ 0.81} & 65.58 $\pm$ 1.03&\underline{75.23$\pm$ 0.29} \\
				& Flights     & 60.72 $\pm$ 1.26 & 61.99 $\pm$ 1.41 & \underline{63.40 $\pm$ 0.29} & 59.66 $\pm$ 1.02 & 56.58 $\pm$ 0.41 & \textbf{63.49 $\pm$ 0.22} & 63.32 $\pm$ 0.16 & 56.05 $\pm$ 0.19& 61.23$\pm$ 0.30\\
				& Contact     & 90.80 $\pm$ 1.19 & 88.87 $\pm$ 0.70 & 93.76 $\pm$ 0.43 & 88.85 $\pm$ 1.37 & 74.79 $\pm$ 0.35 & 90.37 $\pm$ 0.12 & 90.04 $\pm$ 0.27 & \underline{94.14 $\pm$ 0.26}&\textbf{95.47$\pm$ 0.23} \\ \cmidrule(lr{0.5em}){2-11}
				& Avg. Rank   &6.12&7.38&4.75&6.00&7.12&4.62&\underline{3.12}&3.75&\textbf{2.12}\\ \bottomrule
			\end{tabular}
		}
	}
	\label{tab:all4}
\end{table*}